\documentclass{article}

\usepackage{arxiv}

\usepackage[utf8]{inputenc} 
\usepackage[T1]{fontenc}    
\usepackage{hyperref}       
\usepackage{url}            
\usepackage{booktabs}       
\usepackage{amsfonts}       
\usepackage{nicefrac}       
\usepackage{microtype}      
\usepackage{lipsum}
\usepackage[title]{appendix}

\usepackage{xcolor}

\usepackage[linesnumbered,ruled,vlined]{algorithm2e}

\usepackage{stmaryrd}
\usepackage{empheq}
\usepackage{wrapfig}
\usepackage{wrapfig}
\usepackage{multicol}
\usepackage[english]{babel}
\usepackage{blindtext}
\usepackage{tabularx}
\usepackage{pdflscape}
\usepackage{multirow}
\usepackage[inline]{enumitem}
\usepackage{subcaption}

\usepackage{floatrow}
\newfloatcommand{capbtabbox}{table}[][\FBwidth]

\usepackage{natbib}

\SetCommentSty{mycommfont}

\SetKwInput{KwInput}{Input}
\SetKwInput{KwOutput}{Output}
\SetKwInput{KwInit}{Initialization}
\SetKwProg{Fn}{Function}{}{}
\SetKwInput{TM}{Modelisation Procedure}

\title{Unsupervised clustering of series using dynamic programming and neural processes}

\author{
  Karthigan Sinnathamby\thanks{Work done while in Master Data Science at EPFL (Lausanne, Switzerland)}\\
  Visium\\
  Lausanne, Switzerland\\
  \texttt{karthigan.sinnathamby@visium.ch} \\
  \And
Chang-Yu Hou \\
Schlumberger-Doll Research\\
Cambridge, MA\\
\texttt{CHou2@slb.com} \\
     \And
 Lalitha Venkataramanan \\
Schlumberger-Doll Research\\
Cambridge, MA\\
\texttt{LVenkataramanan@slb.com} \\
\And
Vasileios-Marios Gkortsas \\
Schlumberger-Doll Research\\
Cambridge, MA\\
\texttt{VGkortsas@slb.com} \\

\And
François Fleuret \\
  University of Geneva\\
  Geneva, Switzerland\\
  \texttt{francois.fleuret@unige.ch} \\
}

\begin{document}
\maketitle

\begin{abstract}
Following the work of~\cite{dp_1}, we are interested in clustering a given multi-variate series in an unsupervised manner. We would like to segment and cluster the series such that the resulting blocks present in each cluster are coherent with respect to a predefined model structure (e.g. a physics model with a functional form defined by a number of parameters). However, such approach might have its limitation, partly because there may exist multiple models that describe the same data, and partly because the exact model behind the data may not immediately known. Hence, it is useful to establish a general framework that enables the integration of plausible models and also accommodates data-driven approach into one approximated model to assist the clustering task. Hence, in this work, we investigate the use of neural processes to build the approximated model while yielding the same assumptions required by the algorithm presented in~\cite{dp_1}.
\end{abstract}


\section{Introduction}
Clustering groups a set of data-points in such a way that data-points in the same cluster are more similar to each other than to those in other groups. It can be achieved by various algorithms that differ significantly in their understanding of what constitutes a cluster and how to efficiently find them. This task can be achieved by various algorithms (the well-known K-means or spectral clustering but also hierarchical clustering \cite{7100308} or density-based clustering \cite{malzer2019hybrid}) that differ significantly in their understanding of what constitutes a cluster and how to efficiently find them.

The work presented in~\cite{dp_1} has designed an unsupervised clustering algorithm for any series through dynamic programming by putting constraints on the number of cluster, the number of transition as well as the minimal size of any block of points. This algorithm assumes that we can model each cluster by a specific process (cluster characterization) and that given this characterization, we have a function to measure how well a point belongs to this cluster (cluster affiliation). As a result, it clusters a given multi-variate series such that the resultant blocks present in each cluster are coherent with respect to these two assumptions. In other words, data points are clustered together if they can be described using the assumed model with the same parameters.

In~\cite{dp_1}, the authors gave an example use-case by clustering the multi-variate series for an array of petrophysical measurements through a modelisation with a known physics model correlating these data. In practice, such modelisation process can be non-unique as there may exist multiple plausible models describing the same dataset, a scenario often occurring for complex systems. In addition, an underlying model may not readily known. Hence, we are interested in having a more general framework that makes use of deep neural networks to tackle the non-unique modelisation case as well as the scenario for the absence of known models.

In this paper, we augment the clustering framework presented in~\cite{dp_1} with a more general modelisation procedure with neural processes as introduced by~\citet{kim2019attentive}. The neural processes use a combination of neural networks that approximate the underlying processes with a distribution of stochastic models, a concept similar to the stochastic processes. Given blocks of coherent datasets, the learning procedure of neural processes is designed to provide the probabilistic distribution of unknown (targets) points conditioned with a number of coherent data (context) points. Without an explicit functional form for the underlying model within the framework of neural processes, we will demonstrate that one can directly use the context points as the signature of the cluster characterization and use the likelihood of target points as the measure to partition the cluster affiliation. As a result, the approximated models established through neural processes satisfy the two requirements for using the algorithm introduced in \cite{dp_1}.

\section{Neural Process}

\subsection{Description}

We use neural processes (NPs) as described in \citet{garnelo2018neural}. In particular, we make use of the version with attention mechanism: the attentive neural processes (ANPs) introduced in \citet{kim2019attentive}. NPs offer an efficient method to modeling a distribution over regression functions: they can predict the distribution of an arbitrary \textit{target point} conditioned on a set of \textit{context points} of an arbitrary size. This flexibility of NPs enables them to model data that can be interpreted as being generated from a stochastic process. NPs are trained on samples from multiple realizations of a stochastic process (i.e. trained on many different functions). 

Throughout our discussions, we will use the following notations. The variable y designates the output of the regression and x is the input. $\bar{C}$ is the subset of the context points and $\bar{T}$ of the target points.

\begin{figure}[H]
\centering
    \begin{subfigure}[t]{0.49\textwidth}
        \centering
        \includegraphics[width=0.99\linewidth]{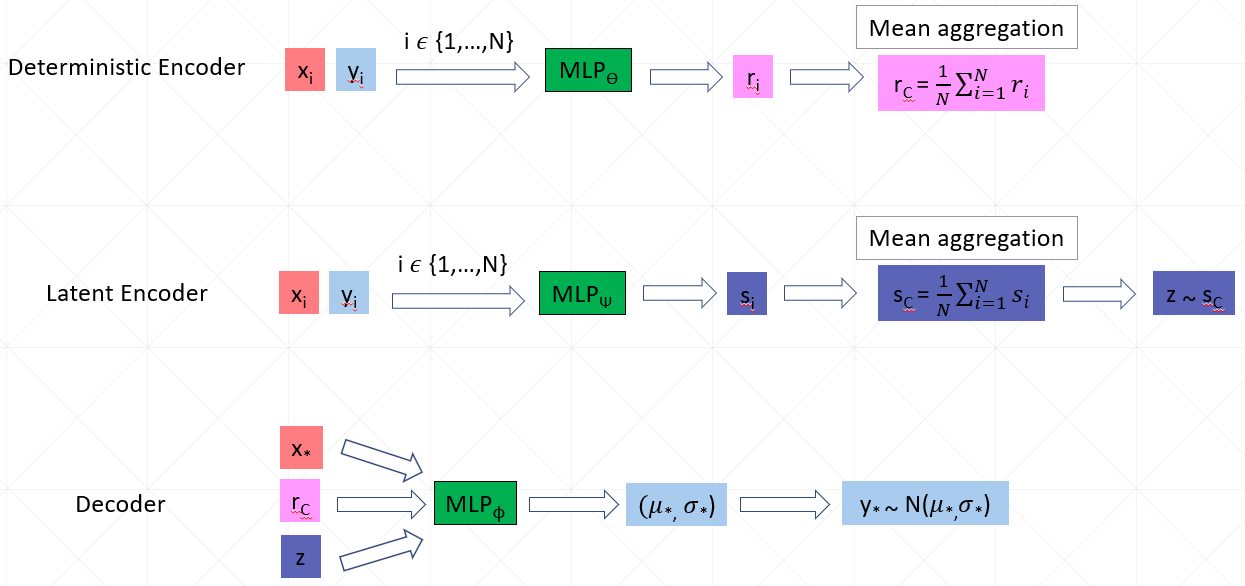}
        \caption{Neural Process}
    \end{subfigure}
    \begin{subfigure}[t]{0.49\textwidth}
        \centering
        \includegraphics[width=0.99\linewidth]{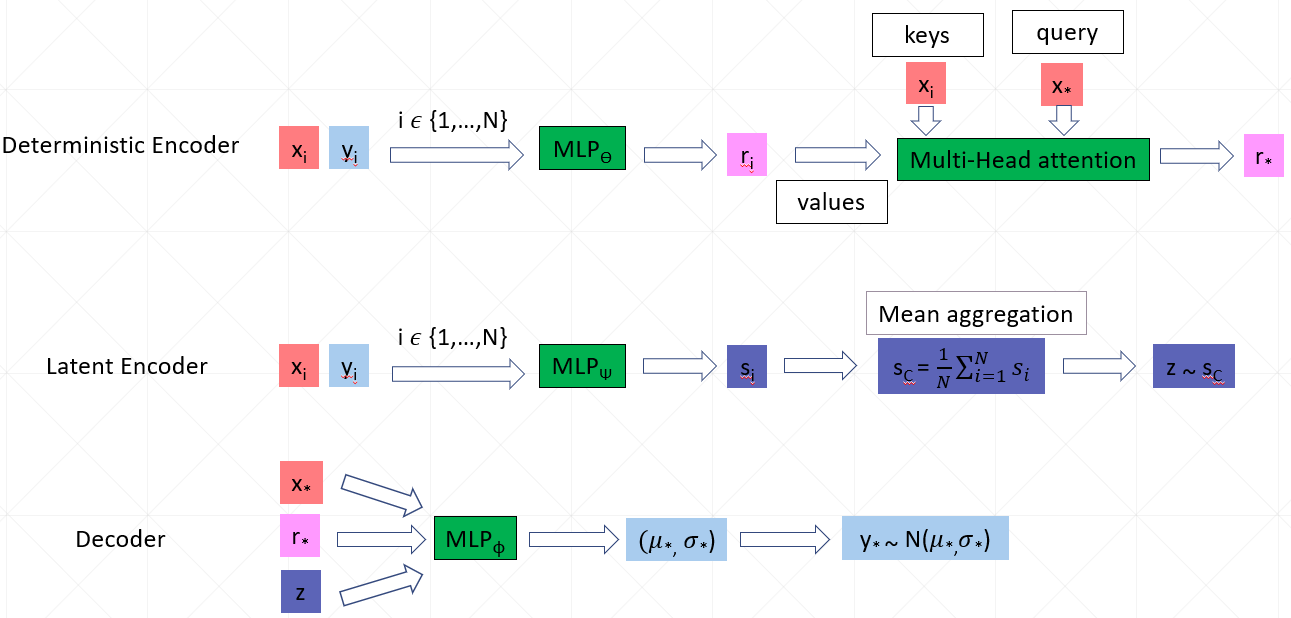}
        \caption{Attentive Neural Process}
    \end{subfigure}
\caption{Model architecture for the NP (left) and ANP (right): in prediction, $x_{i}$ and $y_{i}$ are the context points and $x_{*}$ and $y_{*}$ are the target points to predict. Inspired from figure 2 of \citet{kim2019attentive}.}
\label{fig:anp_np_architecture}
\end{figure}

A NP consists of a set of three neural networks: a latent encoder, a deterministic encoder and a decoder. Each neural network has a main multi-layer perceptron (MLP) part: $MLP_{\psi}$ for the latent encoder, $MLP_{\theta}$ for the deterministic encoder and $MLP_{\phi}$ for the decoder as depicted in figure \ref{fig:anp_np_architecture}. The inputs to the encoders are the points ($x_{i}$,$y_{i}$). The deterministic encoder outputs $r_{i}$ for each pair ($x_{i}$,$y_{i}$) of context points and yields eventually $r_{\bar{C}}$ after applying a mean operation over the $r_{i}$. The latent encoder outputs $s_{i}$ for each pair ($x_{i}$,$y_{i}$) of context points and yields eventually $s_{\bar{C}}$ which parametrizes a Gaussian multivariate distribution, from which a variable $z$ is sampled. If the points ($x_{i}$,$y_{i}$) are the target points, we note $s_{\bar{T}}$ instead of $s_{\bar{C}}$. Finally, the decoder takes as input $r_{\bar{C}}$, $z$ and $x_{*}$ (the input of the target points) and outputs the mean and the standard deviation of $y_{*}$ (the output of the target points) assuming a Gaussian distribution to account for the observation noise. We present the architecture in the figure \ref{fig:anp_np_architecture} which is inspired from figure 2 of \citet{kim2019attentive}.

The parameters of \textit{all the networks are jointly} learned by maximising the following Evidence Lower Bound (ELBO) as in \citet{kim2019attentive}:

\begin{equation}
\label{eqn:loss}
\boxed{\log p(y_{\bar{T}}|x_{\bar{T}},x_{\bar{C}},y_{\bar{C}}) \geq E_{q(z|s_{\bar{T}})}  [\log p(y_{\bar{T}}|x_{\bar{T}},r_{\bar{C}},z)] - D_{KL} ( q(z|s_{\bar{T}}) \Vert q(z|s_{\bar{C}}) )}
\end{equation}

where ($x_{\bar{T}}$,$y_{\bar{T}}$) are the target points, ($x_{\bar{C}}$,$y_{\bar{C}}$) the context points, $p$ is a Gaussian distribution parametrized by the decoder, $q$ is a Gaussian distribution parametrized by the latent encoder.

\subsection{Integration to Dynamic Programming}

The algorithm described in \cite{dp_1} assume that one provides a cluster characterization and a cluster affiliation functions.\textit{ The \textit{context points} represent here the points of a cluster. This is the key link between NPs and the clustering algorithm.} The NP network predicts the output of a target point from its input $x_{*}$, given the input $x_{\bar{C}}$ and output $y_{\bar{C}}$ of the context points. 
As such, $r_{\bar{C}}$ and $z$ (as defined in the figure \ref{fig:anp_np_architecture}) represent the weights of the cluster (cluster characterization). If using attention mechanism, the encoders compute the weights of the cluster by attending to the target point ($r_{*}$). In this case, we do not have an explicit cluster weights but the encoders do compute implicitly a representation of the context points.
As for the cluster affiliation, we can use the NLL part of the loss.

\section{Application to petrophysics}

Here we revisit the petrophysical application for the unsupervised clustering scheme established in~\citet{dp_1}. The difference here is that we first apply the modelisation procedure with the neural processes to learn an approximate model before applying the dynamic programming clustering. The neural processes will be trained with three plausible models describing the physics relations among data points. This is an effective way to integrate multiple models into the stochastic regression process. Here, we will demonstrate how we can effectively use the modelisation of neural processes for the unsupervised clustering based on the dynamic programming.

\subsection{Physics Equations}

Let us start by reviewing the petrophysical problem studied in~\citet{dp_1}. We are interested in clustering the multi-variate series of petrophysical measurements/interpretations consisting of conductivity response of the water-filled rock ($\sigma_{o}$), the porosity ($\phi$), the volume fraction of clay ($f_{clay}$) and the water saturation fraction ($S_{w}$). These measurements are often collected and presented as consecutive data points along the depths of the well, and are often refereed as logs. In~\citet{dp_1}, the authors have used a particular relation between these quantities prescribed by the Waxman-Smits (WS) equation~\cite{Waxman1968}. However, many other physics relations have been established to correlate these quantities. To exhibit how the neural processes can be used to combine multiple models, we will adapt three known physics models described by WS, Sen-Goode-Sibbit (SGS) and Archie's equations, for training an integrated neural process model~\cite{Sen1988}. 

All three considered physics models aim to correlate the aforementioned quantities with a small number of parameters that control the underlying physics processes. Both WS and SGS equations are established to account for the clay effect to the conductivity of water-filled rocks, while Archie's equation is for rocks with negligible clay effects. Specific forms of these equations are listed below.

The WS equation is:
\begin{equation}
	\label{eqn:ws}
	\sigma_{o} = \phi^{m} \times S_{w}^{n} \times \left( \sigma_{w} + \frac{B \times Q_{v}}{S_{w}} \right),
\end{equation}
where $m$ and $n$ are some parameters and $B$ is a function of the temperature $T$ which is considered constant throughout the studied logs. In addition, $\sigma_w$ represents the conductivity of water filling the pore space, and $Q_{v}$ is the volume concentration of clay exchange cations with respect to the pore space, which is associated with the cation exchange capacity (CEC) of the rock:
\begin{equation}
\label{eqn:qv}
Q_{v} = \frac{CEC \times f_{clay} \times (1-\phi)}{\phi} .
\end{equation}

The SGS equation is:
\begin{equation}
\label{eqn:SGS}
\sigma_{o} = \phi^{m}  \times  S_{w}^{n}  \times  \left( \sigma_{w} + \frac{1.93 \times m \times \mu_{T} \times Q_{v}}{1 + \frac{0.7  \times  \mu_{T}  \times  S_{w}^{-n}}{\sigma_{w}}} \right) + 1.3  \times  \mu_{T} \times  \phi^{m}  \times  Q_{v},
\end{equation}
where
\begin{equation}
\label{eqn:mu}
\mu_{T} = 1 + 0.0414  \times  (T-22)
\end{equation}
is the effective mobility of double layer cations with $T$ the temperature given in Celsius. Here,  $m$ and $n$ are parameters similar to those in WS equation, $\sigma_w$ is the conductivity of water filling the pore space and $Q_v$ is defined in Eq.~\eqref{eqn:qv}.

When the effect of clay becomes negligible, namely, $Q_v\to 0$, both WS and SGS equations effectively reduce to the Archie's equation:
\begin{equation}
\label{eqn:archie}
\boxed{\sigma_{o} = \phi^{m}  \times  S_{w}^{n}  \times  \sigma_{w} }
\end{equation}
that describe the rock conductivity response of clean formations in the absence of clay. Although one may think that Archie's equation is simply a special case for WS and SGS equations, it is important to include this limiting case as an independent relation for the training of NPs, not only because it represents a huge class of rocks, but also because it provide a relation directly discount the clay effect.  

Now, we have three equations relating the interested logs. These equations with a specific set of parameters can be seen as the realization of a stochastic process. A neural process trained on data generated by these three equations would consist of a more general modelisation or at least an aggregated one. Instead of having explicit parameters to define a coherent cluster, one would directly use the \textit{context points} (the cluster) as the representation of a cluster that define the regression function over the distribution approximated by the NP.

\subsection{Neural Process Training}

Here we provide some details on our training procedure to integrate three physics models into the neural processes.

Let us start by describing how we build/generate the dataset. We will fix all the parameters controlling the underlying physics processes, $m$, $n$, $\sigma_w$ and/or CEC, within a physically realistic range. At each epoch, we draw randomly $3000$ (size of the training set) sets of parameters and a physics equation among the three models: each of the $3000$ draws have a randomly chosen set of parameters and a randomly selected equation.

For each set, we first randomly generate three lists of $\phi$, $S_w$ and $f_{clay}$ within a physically realistic range with $200$ elements in each list. For NPs training, these lists are the input of the regression, x. From these three lists and the selected equation from Eq.~\eqref{eqn:ws}--~\eqref{eqn:archie}, we then generate a list of $200$ points of conductivity of rocks with reasonable noise added into the both input parameters and also output signals. For NPs training, this list of conductivities is basically the output of regression, y. Among $200$ quadruples from these lists, between $50$ and $100$ quadruples are used as context points (randomly chosen). The target points consist of all the 200 points. As a result, the training is done with \textit{coherent} context points: we know that they come from a same equation using the same parameters due to the simulation. The target points are also coherent with the context points. In other words, these points should come from the same stochastic process realization.


\begin{figure}
	\centering
	\begin{subfigure}[t]{0.49\textwidth}
		\centering
		\includegraphics[width=0.99\linewidth]{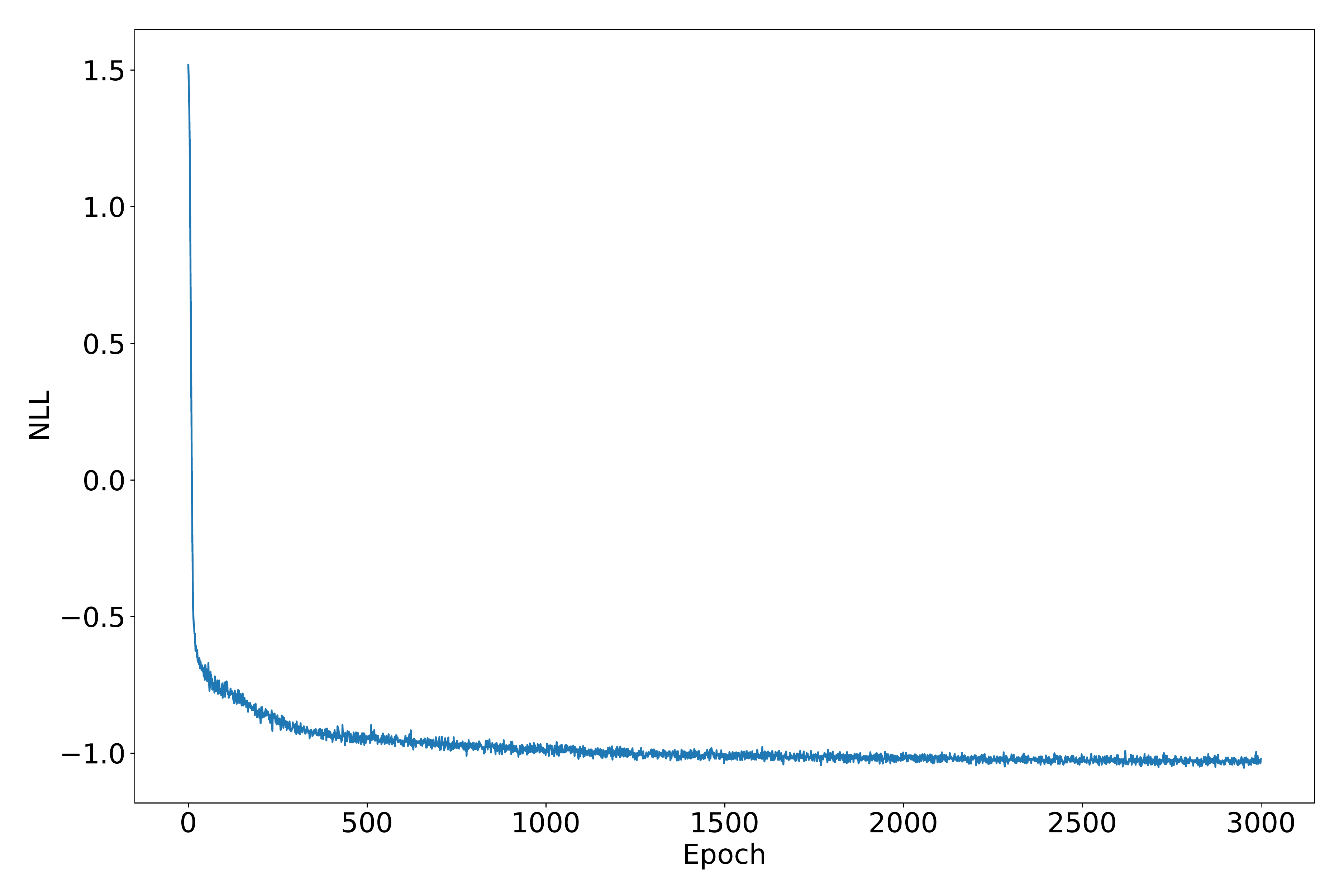}
		\caption{Training loss}
	\end{subfigure}
	\begin{subfigure}[t]{0.49\textwidth}
		\centering
		\includegraphics[width=0.99\linewidth]{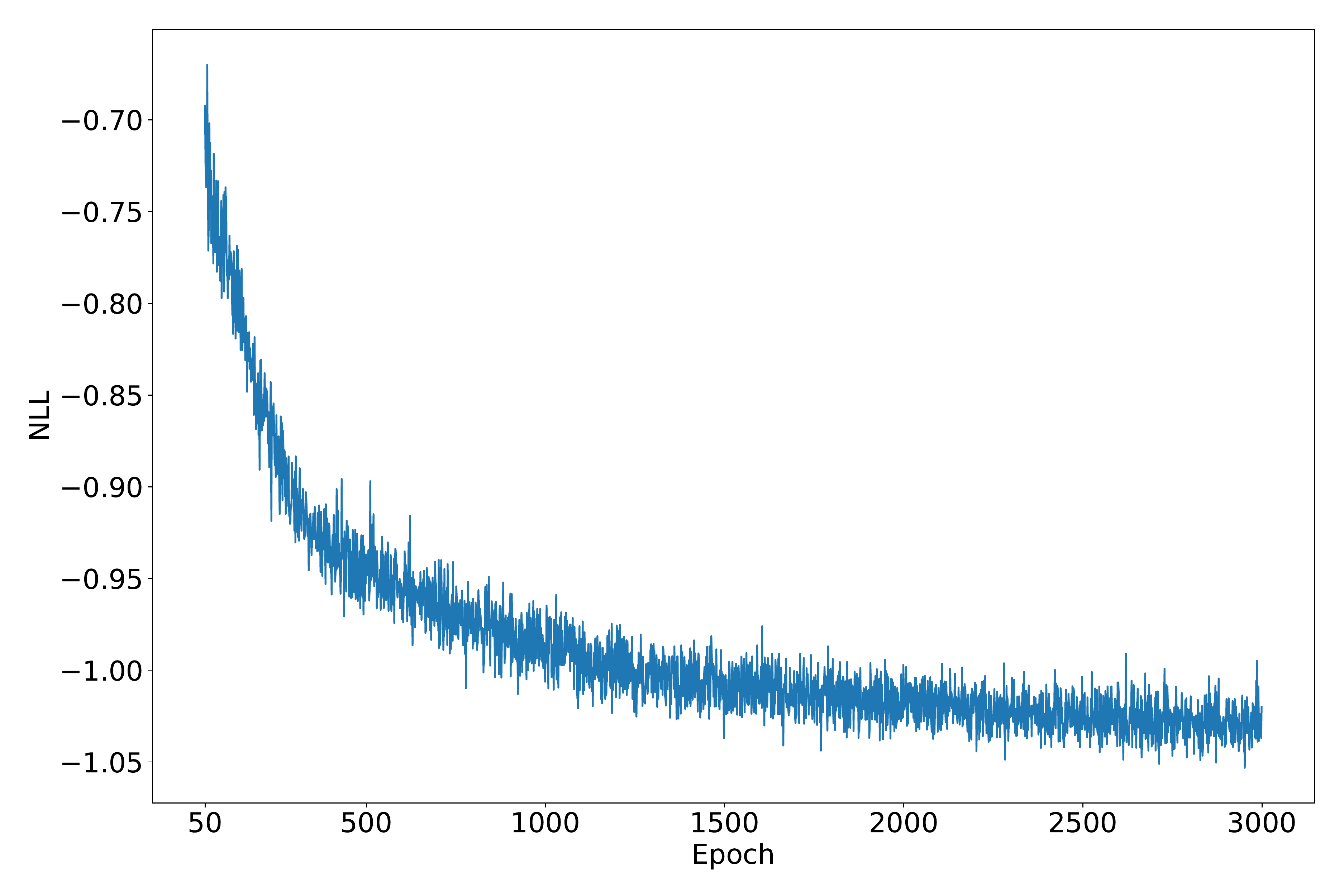}
		\caption{Training loss (starting epoch 50)}
	\end{subfigure} 
	\caption{Training of ANP. \textit{The y-axis is the negative log-likelihood (NLL) and the x-axis is the number of epoch. We give the training loss curve for the training of the ANP (a) and the same curve but starting at epoch 50 (b). \textbf{Beware the scale of y is different between the figures.} }}
	\label{fig:training_loss}
\end{figure}

Eventually, we have used the same structure of ReLU activated linear layers for the three networks as in \citet{kim2019attentive} as well as the same particularization tricks and optimizer. More precisely, we have:
\begin{itemize}
\item the latent encoder consists of three layers (uniform self-attention) before the layer for the mean and the one for the sigma of the Gaussian distribution,
\item the deterministic encoder has two layers (uniform self-attention) and a cross-attention module with 8 heads (the keys and queries are first passed through two layers),
\item the decoder has three layers.
\end{itemize}
All layers have a hidden output size of 16 (except the last one of the decoder which has two - the mean and sigma of the one-dimensional regressed value), giving up 3778 parameters in total. We give the training Negative log likelihood (NLL) curve in Fig. \ref{fig:training_loss}.

\section{Experiments \& Results}

We have conducted the same experiments as in~\cite{dp_1}, where multiple examples of synthetic multi-variate series of $\sigma_{0}$, $\phi$, $S_w$ and $f_{clay}$ are generated with multiple blocks belonging to different clusters. All the data points have noise added to them in the level expected from measurement and interpretation precision. In total, seven testing examples are generated and used to test the dynamic programming clustering algorithm together with the NPs models. Five examples consist of blocks satisfying the either WS or Archie's equation with varying parameters (WS-1, WS-2, WS-2-smooth, WS-3 and WS-3-smooth). Two examples haves blocks obeying WS, SGS or Archie's equation (SGS-1 and SGS-2). 

Similar to the strategy discussed in~\cite{dp_1}, we perform the grid search by applying the algorithm with a range of the number of cluster, $C_{grid}$, and the number of transition, $N_{grid}$. After the grid search, we present the final clustering results of the lowest NLL and of the most common pattern among a selected combinations of $(C_{grid}, N_{grid})$. Table~\ref{tab:results_ANP} summarizes the cost (NLL) of the clustering pattern and Adjust Rand Index (ARI) between the clustering results and the ground truth of the generated examples. For comparison, we also list the cost (L2 loss) and ARI of the clustering results using the same algorithm and WS equation as the underlying physics equation.

\begin{table}
	\begin{center}
		\begin{tabular}{||c c c c c c||} 
			\hline
			Datasets & Method & Prediction & $cost_{pred}$ & $cost_{true}$ & ARI  \\ [0.5ex] 
			\hline\hline
			
			WS-1 & WS & Most common &  0.170 & 0.170 & 0.76 \\ 
			& \textbf{ANP} & Most common & -1.06 & -1.11 & 0.83 \\ 
			&  & \textbf{Lowest cost} & \textbf{-1.12} & \textbf{-1.11} & \textbf{0.96} \\ 
			
			\hline
			WS-2 & WS & Most common & 0.161 & 0.165 & 0.91 \\ 
			& \textbf{ANP} & Most common & -1.07 & -1.09 & 0.74 \\ 
			&  & \textbf{Lowest cost} & \textbf{-1.09} & \textbf{-1.09} & \textbf{0.99} \\ 
			
			\hline
			WS-2-smooth & \textbf{WS} & \textbf{Most common} & \textbf{0.164} & \textbf{0.161} & \textbf{0.82} \\ 
			& ANP & Most common & -1.07 & -1.09 & 0.70 \\   
			&  & Lowest cost & -1.07 & -1.09 & 0.70 \\ 
			
			\hline
			WS-3 & WS & Most common &  0.170 & 0.166 & 0.81 \\ 
			& \textbf{ANP} & Most common & -1.10 & -1.13 & 0.93 \\ 
			&  & \textbf{Lowest cost} & \textbf{-1.13} & \textbf{-1.13} & \textbf{0.98} \\ 
			
			\hline 
			WS-3-smooth & WS & Lowest cost & 0.166 & 0.164 & 0.84 \\ 
			& \textbf{ANP} & Most common & -1.08 & -1.13 & 0.86 \\
			&  & \textbf{Lowest cost} & \textbf{-1.13} & \textbf{-1.13} & \textbf{0.98} \\ 
			
			\hline
			SGS-1 & \textbf{WS} & \textbf{Most common} & \textbf{0.163} & \textbf{0.161} & \textbf{1.0} \\ 
			& ANP & Most common & -1.08 & -1.10 & 0.80 \\ 
			&  & Lowest cost & -1.12 & -1.1 & 0.94 \\ 
			
			\hline
			SGS-2 & WS & Most common & 0.166 & 0.165 & 0.82 \\ 
			& \textbf{ANP} & Most common & -1.10 & -1.14 & 0.86 \\
			&  & \textbf{Lowest cost} & \textbf{-1.15} & \textbf{-1.14} & \textbf{1.00} \\ 
			
			\hline
		\end{tabular}
	\end{center}
	\caption{Results for simulated data. \textit{For each dataset, we have a line for the best answer using WS (same results as in \cite{dp_1}) and a line for the most common and the lowest cost answers using ANP. $cost_{pred}$ is the cost of the prediction and $cost_{true}$ is the cost of the ground truth answer. \textbf{Beware, the cost for the WS is the MSE while for ANP it is the NLL.} ARI is the Adjusted Rand Index between the prediction and the ground truth as defined in \cite{dp_1}. In bold, we have the prediction with the highest ARI.}}
	\label{tab:results_ANP}
\end{table}

From the ARI scores in table \ref{tab:results_ANP}, the clustering using the method with NP provides very reliable predictions over all the testing datasets. For WS-1, WS-2, WS-3, WS-3-smooth and SGS-2, the lowest cost answer of the ANP yields a better ARI than that in the WS case. Moreover, the most common pattern of the ANP also yields a comparable answer compared to the clustering pattern using WS equation as characterization reference. Even though it may not represent the most generic case, a pleasant surprise is that the lowest cost pattern of the ANP not only yields a better answer but it also gives almost a perfect pattern as indicated by their ARI scores.

\begin{figure}
	\begin{floatrow}
		\ffigbox{%
			\includegraphics[trim=2cm 1cm 4cm 2cm,clip,scale=0.35]{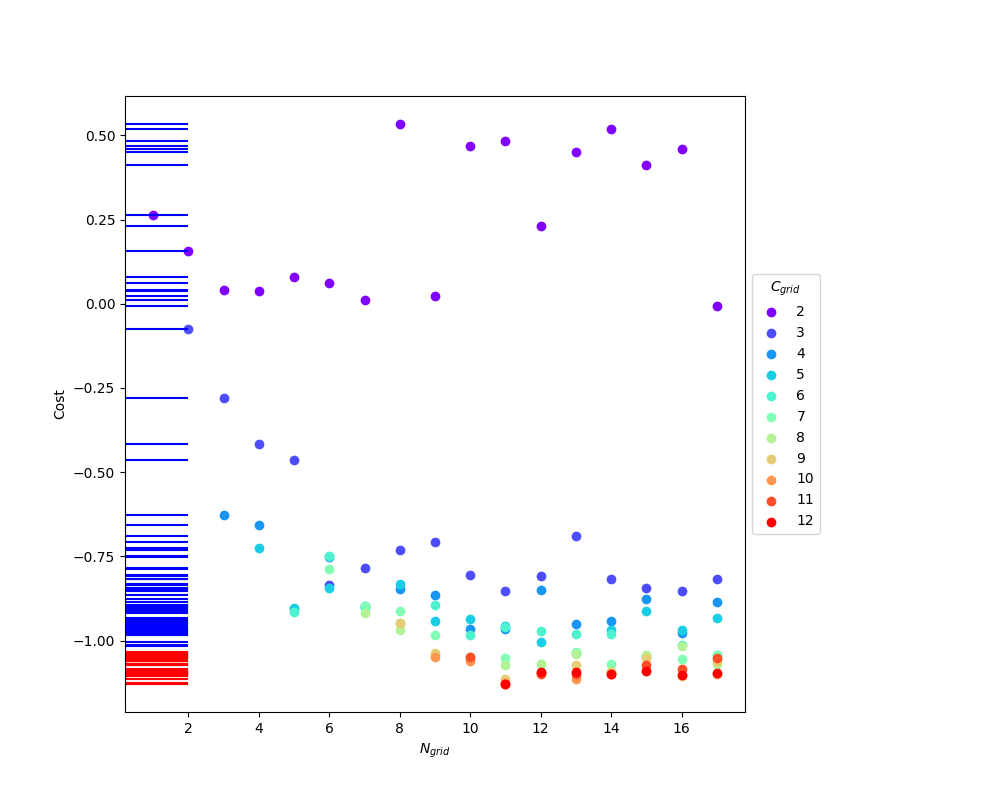}
		}{%
			\caption{Grid-search criterion for WS-3\\ Most common: $N_{grid}=8$, $C_{grid}=6$}
			\label{fig:grid-WS-3}%
		}
		\capbtabbox{%
			
			\begin{tabular}{||c c c c c||} 
				\hline
				label & m & n & $\rho_{w}$ & CEC \\ [0.5ex] 
				\hline\hline
				0 & 1.85 & 1.7 & 0.03 & 0 \\ 
				\hline
				1 & 2.0 & 2.0 & 0.03 & 0 \\
				\hline
				2 & 2.05 & 2.0 & 0.029 & 30 \\
				\hline
				3 & 2.3 & 2.1 & 0.031 & 0 \\
				\hline
				4 & 2.5 & 2.2 & 0.049 & 80 \\
				\hline
				5 & 2.0 & 2.5 & 0.05 & 0 \\
				\hline
				6 & 2.0 & 1.9 & 0.05 & 0 \\
				\hline
				7 & 2.1 & 2.1 & 0.051 & 45 \\ 
				\hline
			\end{tabular}
			
		}{%
			\caption{Parameters for WS-3}
			\label{table:WS-3}%
		}
	\end{floatrow}
\end{figure}

We will present our clustering results in the following fashion. For each dataset, we give the table of parameters that were used to generate the data along with the cluster label as shown in Table~\ref{table:WS-3}. We then provide a scatter plot of the cost against $N$ for given numbers of clusters as depicted in Fig.~\ref{fig:grid-WS-3}. From this plot, we select the combinations of $(C_{grid},N_{grid})$, indicated by red ticks on the vertical axis, to be used for identifying the most common pattern. Here, we practice the similar selecting criterion as laid out in~\cite{dp_1}.

\begin{figure}[h!]
	\centering
	\begin{subfigure}[b]{0.49\textwidth}
		\includegraphics[trim=4cm 2cm 2cm 2cm,clip,width=1\linewidth]{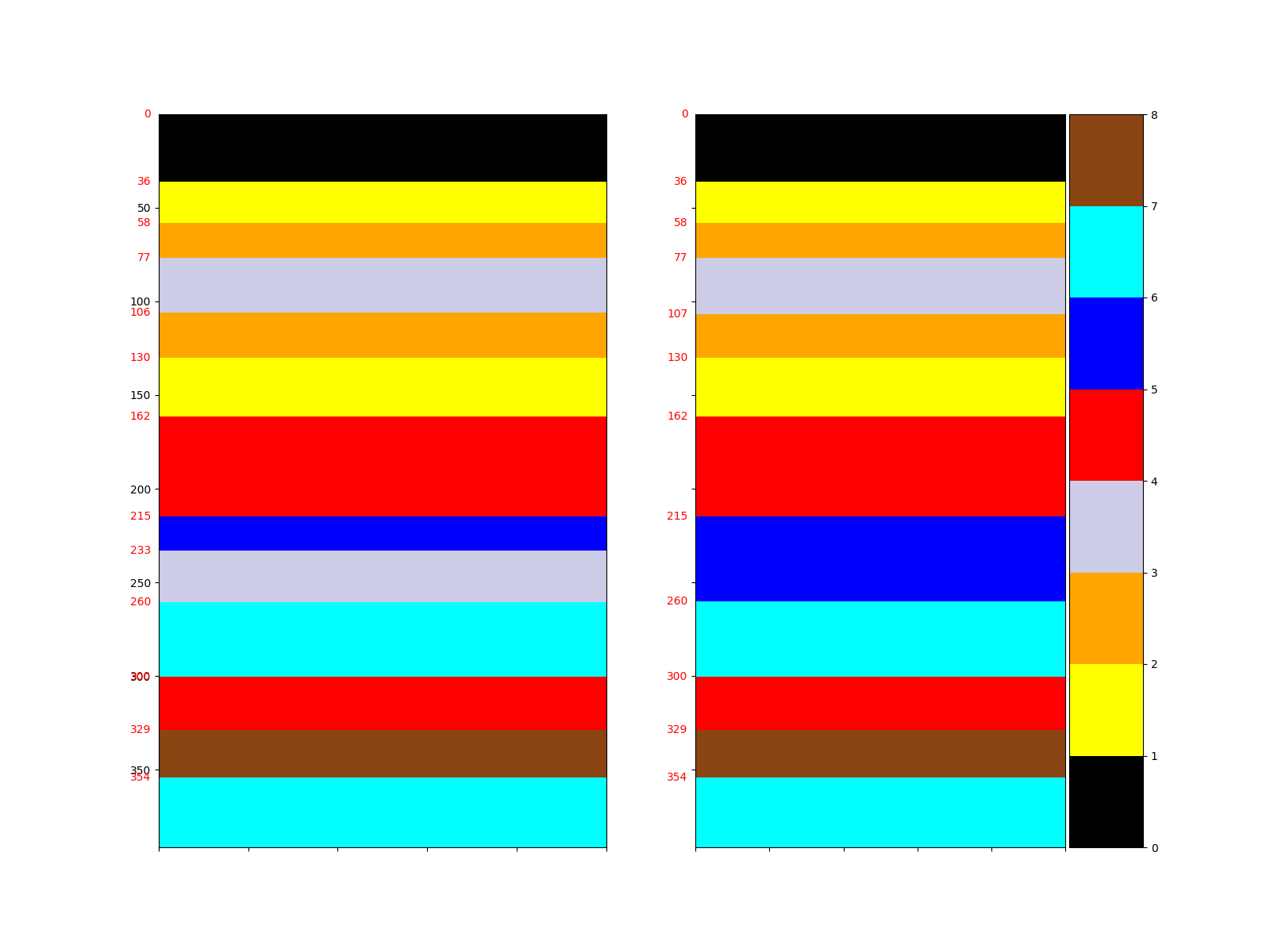}
		\caption{Most common}
	\end{subfigure}
	\begin{subfigure}[b]{0.49\textwidth}
		\includegraphics[trim=4cm 2cm 2cm 2cm,clip,width=1\linewidth]{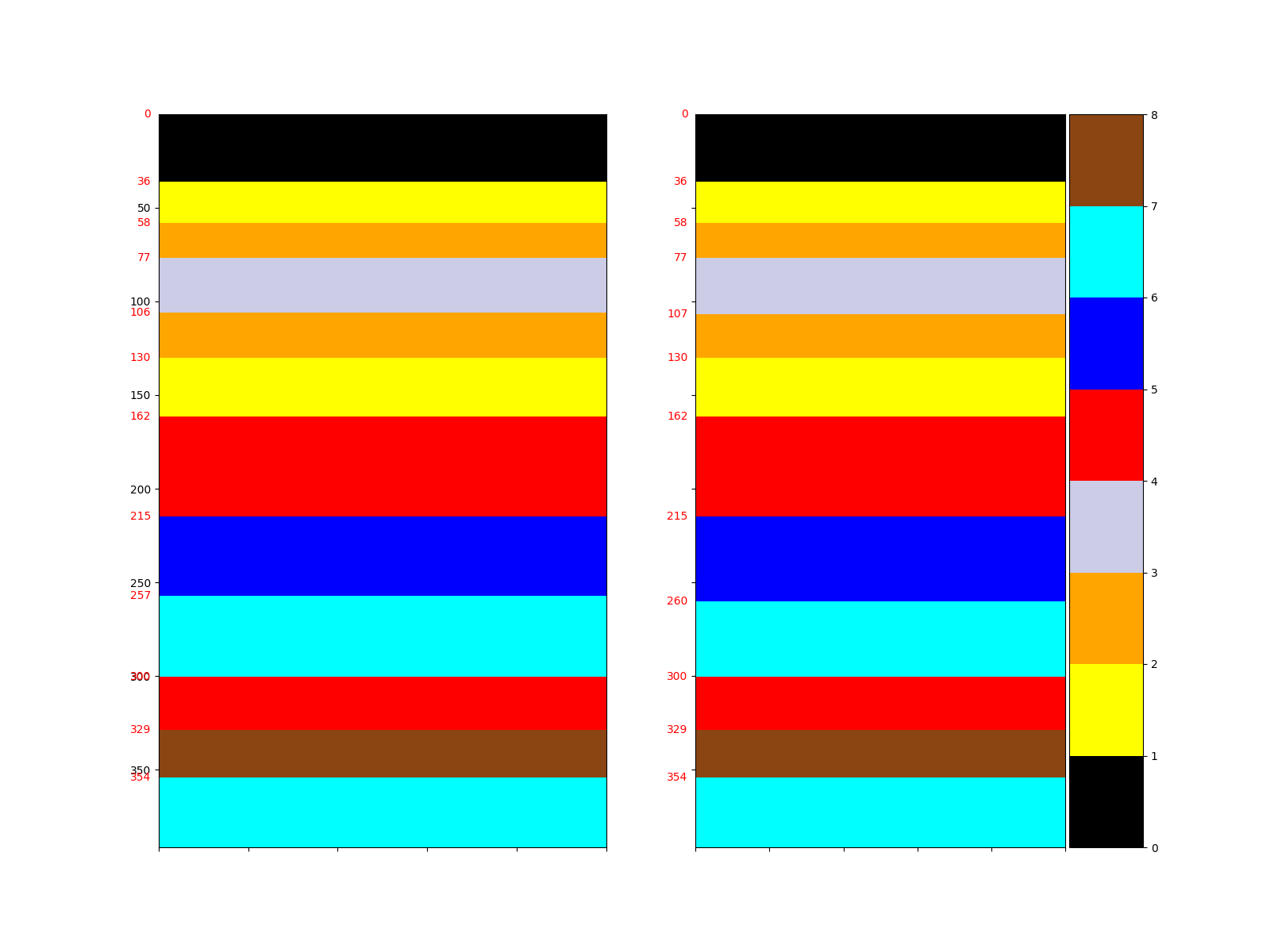}
		\caption{Lowest cost}
	\end{subfigure}
	\caption{Predictions for WS-3}
	\label{fig:WS-3-prediction}
\end{figure}

The clustering predictions from the algorithm with NPs model are shown in plots similar to those in Fig.~\ref{fig:WS-3-prediction} (a) and (b). On those plots, the left figure is either the most common or the lowest cost prediction and the right one represents the ground truth. Each color represents a different cluster. However, the color coding for the ground truth pattern and for the predicted clustering might have nothing in common. The y axis represents the depth and the x axis is simply for illustration purpose.

\begin{figure}[h!]
	\centering
	\begin{subfigure}[b]{0.49\textwidth}
		\includegraphics[trim=0cm 0cm 0cm 2cm,width=1\linewidth]{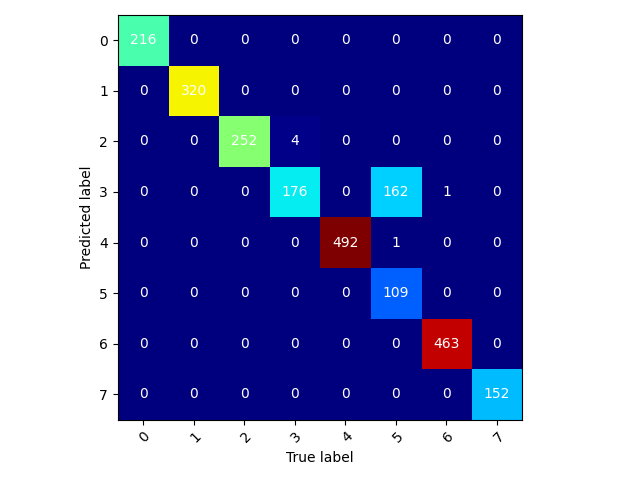}
		\caption{Most common}
	\end{subfigure}
	\begin{subfigure}[b]{0.49\textwidth}
		\includegraphics[trim=0cm 0cm 0cm 2cm,width=1\linewidth]{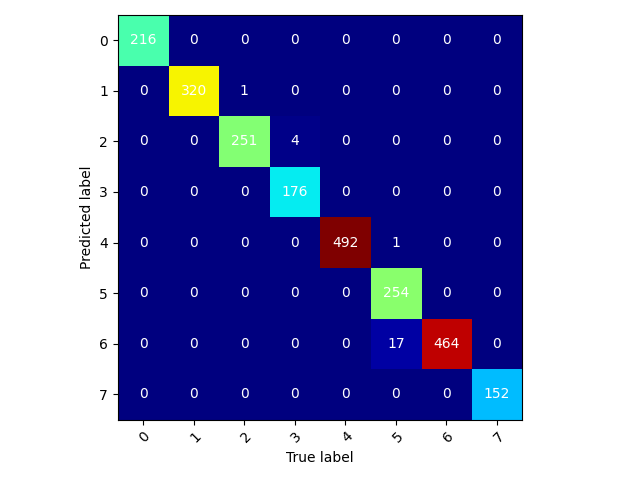}
		\caption{Lowest cost}
	\end{subfigure}
	\caption{Confusion matrices for WS-3}
	\label{fig:WS-3-confusion}
\end{figure}
  
We also give their confusion matrix with respective to the ground truth as shown in Fig.~\ref{fig:WS-3-confusion}. The confusion matrix give us a convenient way to inspect the difference between the cluster predicted by the algorithm and the ground truth. By cross checking the prediction plots and the confusion matrix, we can better judge the quality of the clustering. As shown in Figs.~\ref{fig:WS-3-prediction} and~\ref{fig:WS-3-confusion}, both the most common case and the lowest cost cases show promising clustering results. Here we observe nearly perfect clustering pattern by the lowest cost pattern given in Fig.~\ref{fig:WS-3-prediction} (b) and Fig.~\ref{fig:WS-3-confusion} (b). However, beside a mix-up between the cluster 3 and 5, the most common pattern also provide a quality clustering results as shown in Fig.~\ref{fig:WS-3-confusion}.   

The clustering results for all other testing examples are presented in appendix in similar fashions. We observe similar clustering quality throughout our testing examples, which demonstrate the potential of using a trained NP model as the reference for clustering characteristic.

\section{Discussion}

Compared with the clustering results presented in~\cite{dp_1}, we see much more heterogeneity in the cost scatter plot when using NPs as the underlying characteristic model. Unlike the case directly using WS equation as the reference, where we can see a more horizontal behavior, i.e. for many ($N$,$C$) tuples we have the same or very similar cost, we do not have such a clear behavior on the scatter plots of the NPs method.

This heterogeneity in the cost can be explained by several reasons. First, one must note that we use the NLL from NPs as the cluster affiliation metric, which takes into account both the mean and the variance of the predicted value. This in principle makes the NLL from NPs a more precise measure than the L2 loss from a variance-free physics model such as WS equation. However, when going for bigger $N$ and $C$, the random initialized blocks will be smaller, yielding eventually a fragmented structure which inherently gives a higher variance and hence a higher NLL when predicting the rest of the points given each of these blocks. This is a clear contrast from using a covariance-free model. 

On the other hand, the algorithm might be stuck in a local minimum afterwards providing still low NLL (the NPs are good at their job) but not the lowest achievable NLL. We expect that as $N$ and $C$ grow, the NLL goes down until a plateau is reached. We might not see this phenomenon with the NPs method. Taking into account these remarks, we believe that for pairs ($N$,$C$) with small $N$ and $C$, the method with WS in \cite{dp_1} will have similar behavior as with NPs. However, for higher $N$ and $C$, the behavior is different and more disparate in the NPs case.

We have observed that the lowest cost (NLL) answer in the NPs method might be the best prediction that one can achieve, but we have not been able to prove it. We believe that one must look eventually both at the most common and the lowest cost answer, and decide according to external information such as the approximate values for $N$ and $C$ we expect or other logs data.

Overall the NPs method provide equal or better results than with directly the equations and offer a way to represent the equations in a aggregated way while using the clustering algorithm.

\section{Conclusion}
We have extended the unsupervised clustering algorithm provided in~\cite{dp_1} by using neural processes to approximate the underlying model used to characterize the coherence of the clusters. This enables the user to come up with more expressiveness and compactness using deep neural networks while also providing more reliable answers.

\bibliographystyle{unsrtnat}
\bibliography{references}

\begin{appendices}
\section{Other Datasets}

Here we present the clustering results for all the testing examples, beside WS-3, in the same fashion indicated in the main text.
\newpage

\begin{figure}
\begin{floatrow}
\ffigbox{%
 \includegraphics[trim=2cm 1cm 4cm 2cm,clip,scale=0.3]{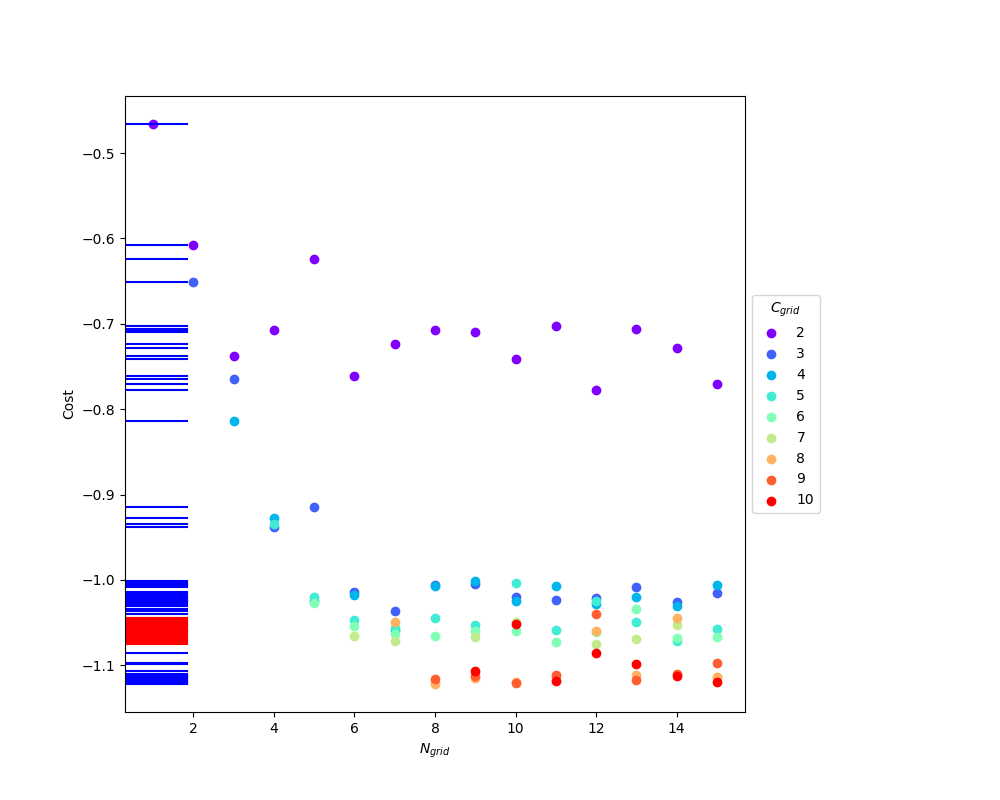}
}{%
  \caption{Grid-search criterion for WS-1 \\ Most common: $N_{grid}=14$, $C_{grid}=10$}%
}
\capbtabbox{%

 \begin{tabular}{||c c c c c||} 
 \hline
 label & m & n & $\rho_{w}$ & CEC \\ [0.5ex] 
 \hline\hline
 0 & 2.1 & 2.3 & 0.052 & 0 \\ 
 \hline
 1 & 1.9 & 1.8 & 0.05 & 0 \\
 \hline
 2 & 1.8 & 1.75 & 0.052 & 0 \\
 \hline
 3 & 2.05 & 1.9 & 0.05 & 0 \\
 \hline
 4 & 2.2 & 2.0 & 0.048 & 20 \\
 \hline
 5 & 2.4 & 2.1 & 0.048 & 60 \\
 \hline
\end{tabular}

}{%
  \caption{Parameters for WS-1}%
}
\end{floatrow}
\end{figure}

\begin{figure}
\centering
  \begin{subfigure}[b]{0.49\textwidth}
    \includegraphics[trim=4cm 2cm 2cm 2cm,clip,width=1\linewidth]{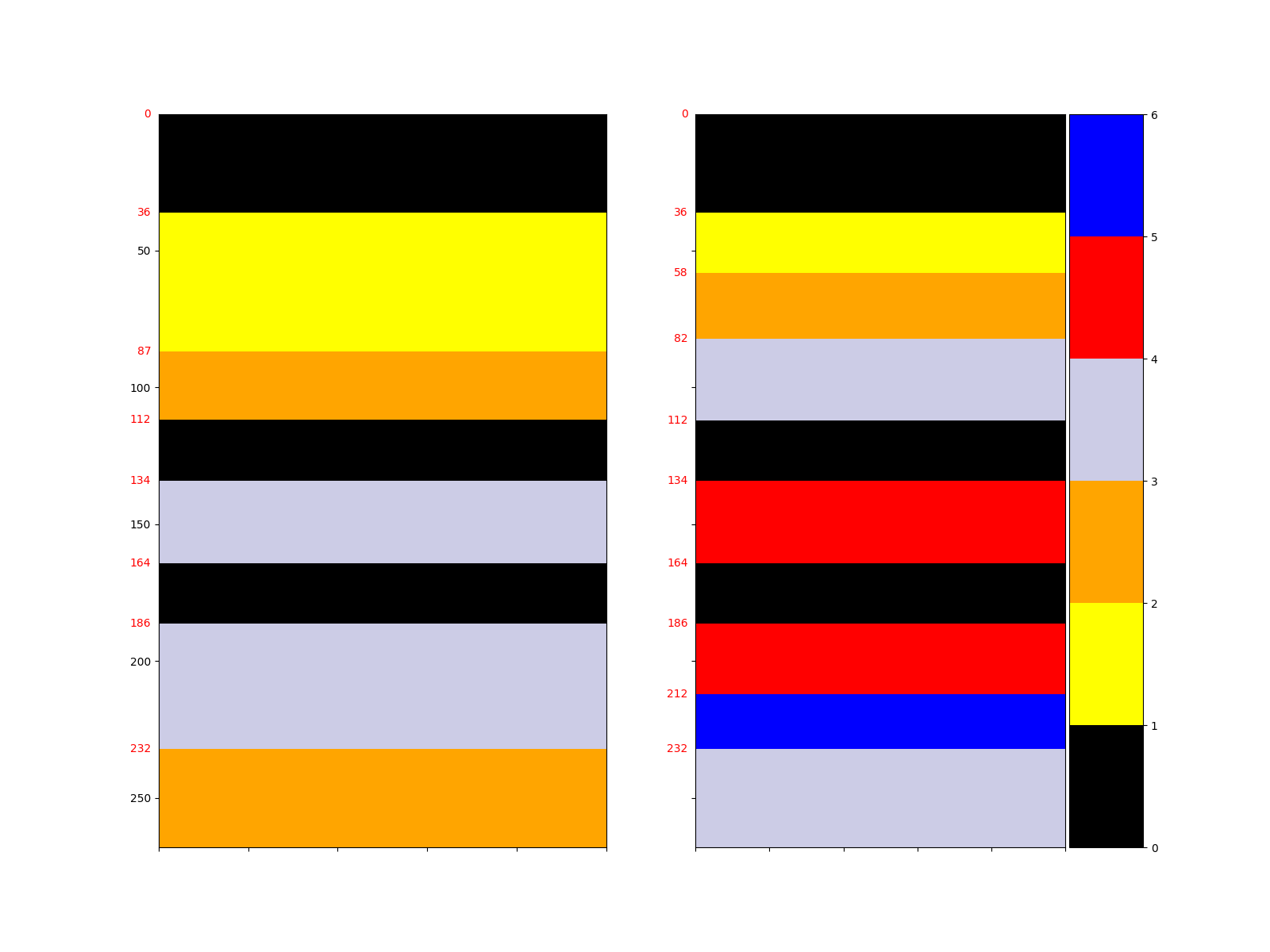}
    \caption{Most common}
  \end{subfigure}
  \begin{subfigure}[b]{0.49\textwidth}
    \includegraphics[trim=4cm 2cm 2cm 2cm,clip,width=1\linewidth]{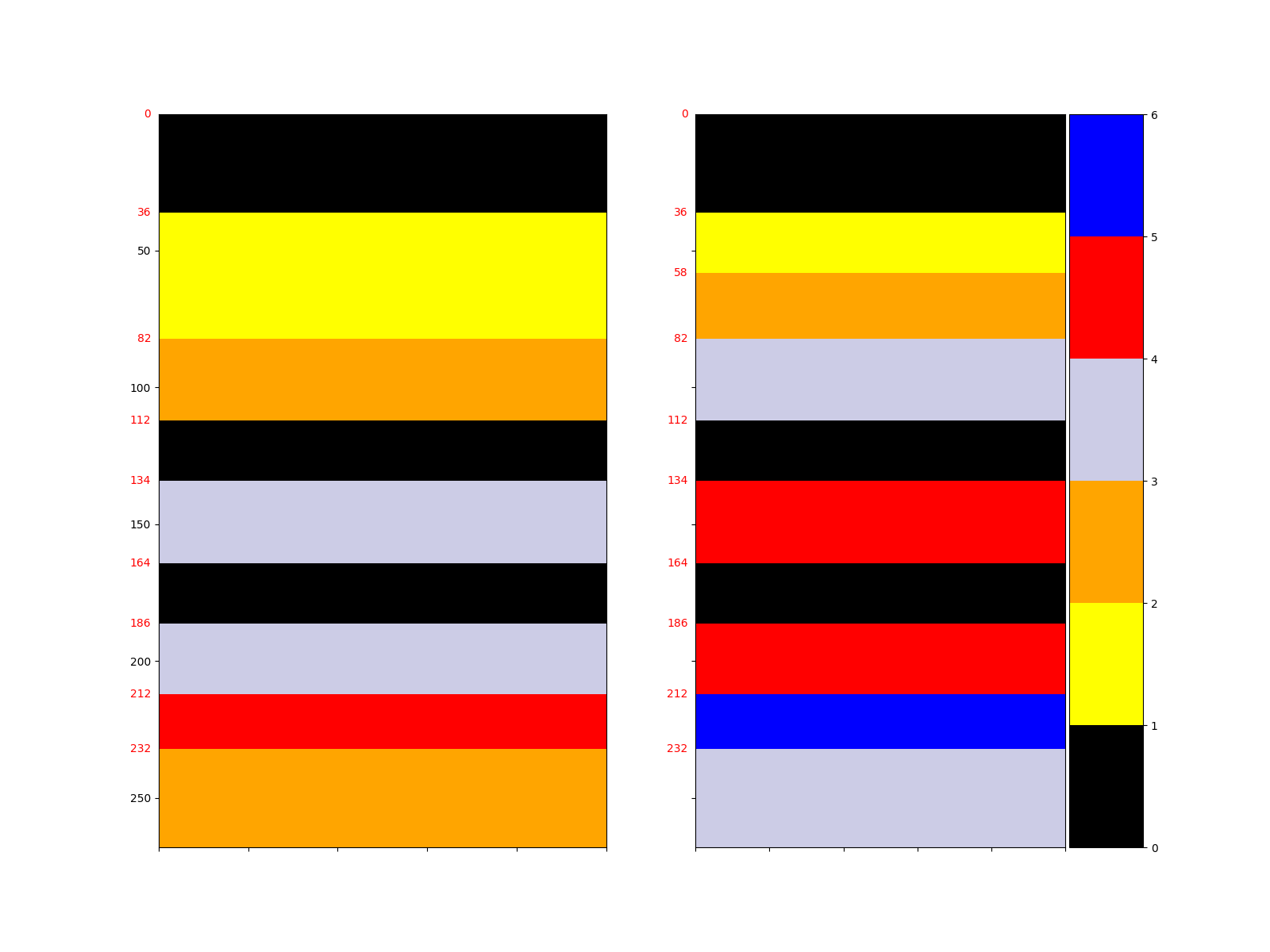}
    \caption{Lowest cost}
  \end{subfigure}
  \caption{Predictions for WS-1}
\end{figure}

\begin{figure}
\centering
  \begin{subfigure}[b]{0.49\textwidth}
    \includegraphics[trim=0cm 0cm 0cm 2cm,width=1\linewidth]{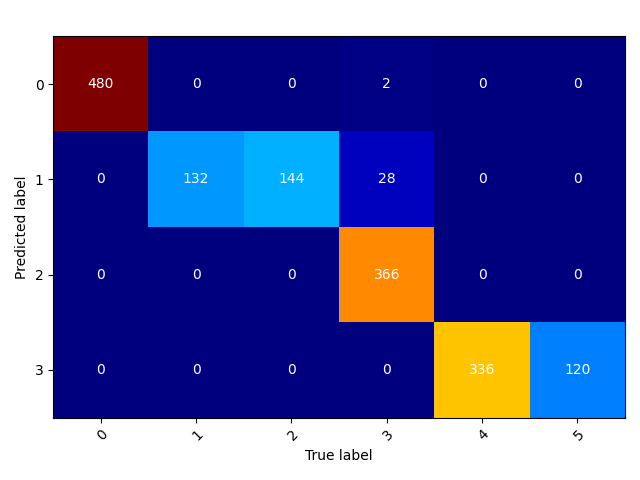}
    \caption{Most common}
  \end{subfigure}
  \begin{subfigure}[b]{0.49\textwidth}
    \includegraphics[trim=0cm 0cm 0cm 2cm,width=1\linewidth]{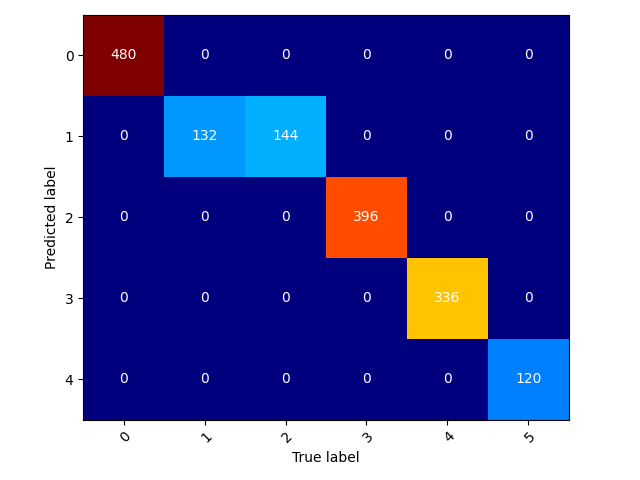}
    \caption{Lowest cost}
  \end{subfigure}
  \caption{Confusion matrices for WS-1}
\end{figure}

\begin{figure}
\begin{floatrow}
\ffigbox{%
        \includegraphics[trim=2cm 1cm 4cm 2cm,clip,scale=0.30]{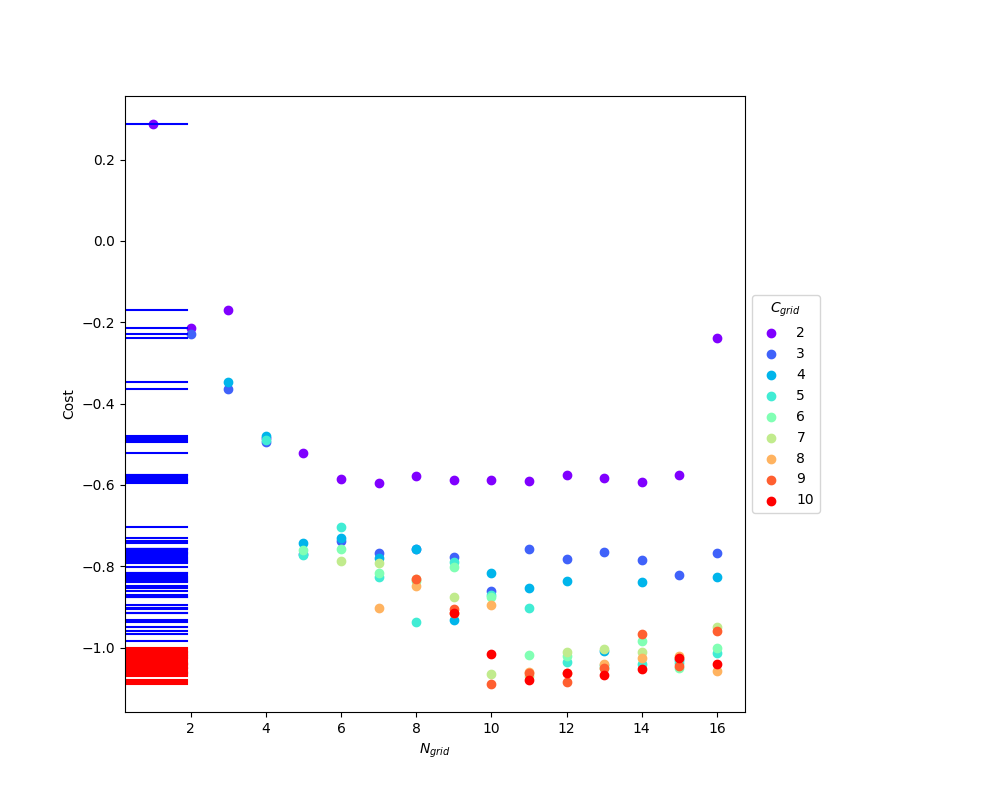}
}{%
   \caption{Grid-search criterion for WS-2 \\ Most common: $N_{grid}=14$, $C_{grid}=8$}%
}
\capbtabbox{%

 \begin{tabular}{||c c c c c||} 
 \hline
 label & m & n & $\rho_{w}$ & CEC \\ [0.5ex] 
 \hline\hline
 0 & 1.85 & 1.8 & 0.052 & 0 \\ 
 \hline
 1 & 2.1 & 2.0 & 0.052 & 0 \\
 \hline
 2 & 2.4 & 2.3 & 0.049 & 80 \\
 \hline
 3 & 1.9 & 2.0 & 0.051 & 0 \\
 \hline
 4 & 2.0 & 1.95 & 0.051 & 30 \\
 \hline
 5 & 2.0 & 2.5 & 0.05 & 0 \\ [1ex] 
 \hline
\end{tabular}

}{%
  \caption{Parameters for WS-2}%
}
\end{floatrow}
\end{figure}

\begin{figure}[h!]
\centering
  \begin{subfigure}[b]{0.49\textwidth}
    \includegraphics[trim=4cm 2cm 2cm 2cm,clip,width=1\linewidth]{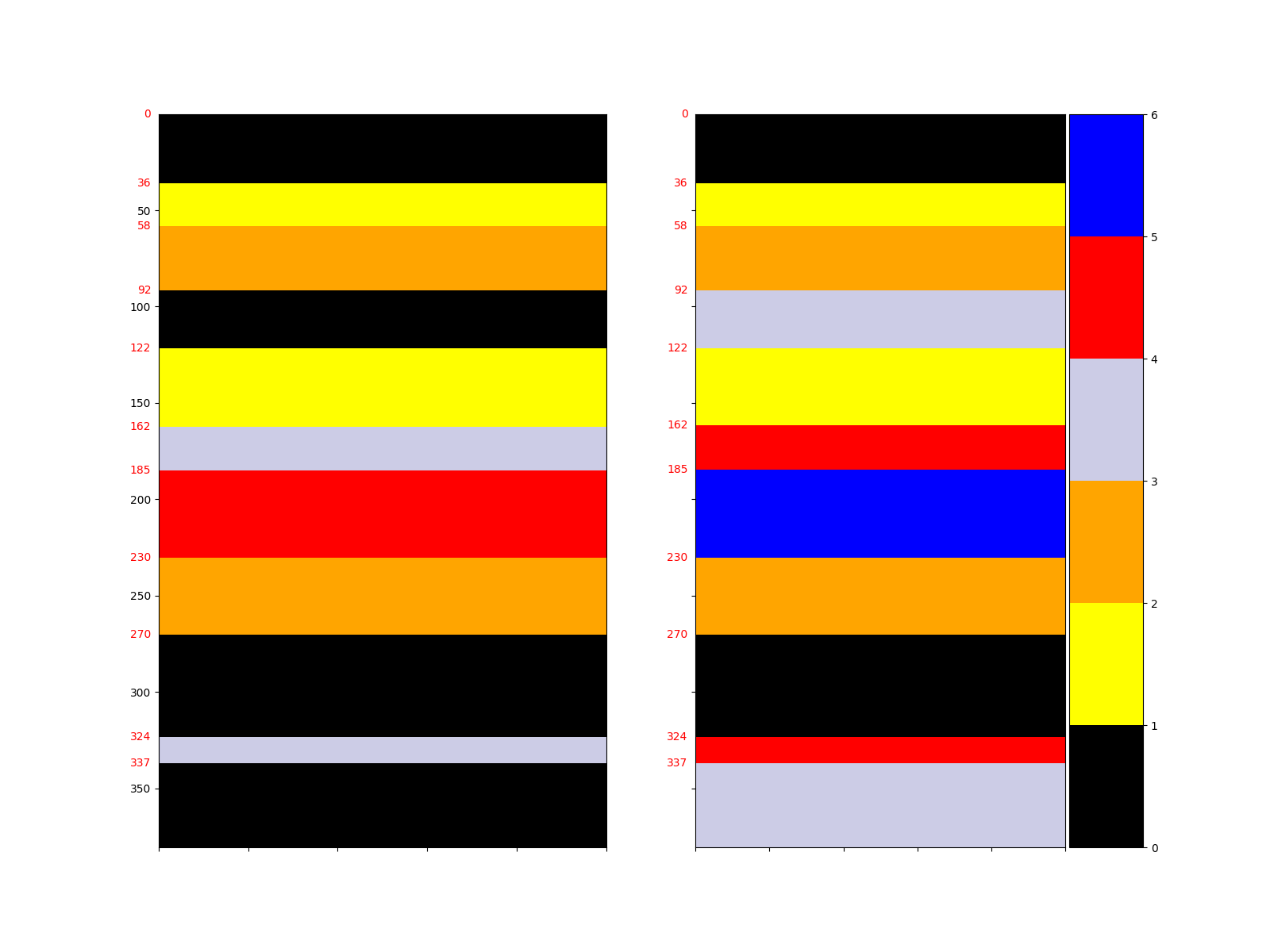}
    \caption{Most common}
  \end{subfigure}
  \begin{subfigure}[b]{0.49\textwidth}
    \includegraphics[trim=4cm 2cm 2cm 2cm,clip,width=1\linewidth]{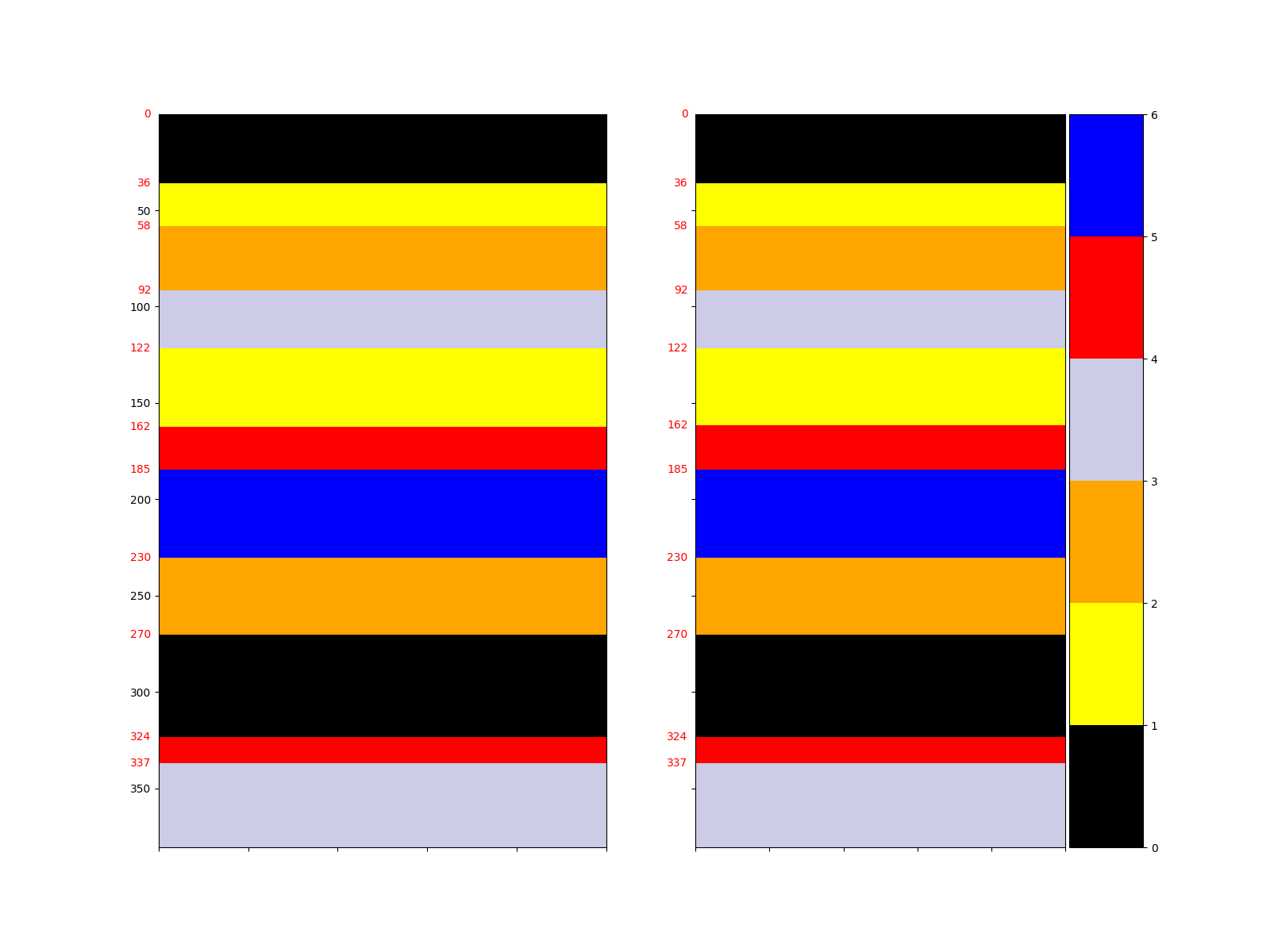}
    \caption{Lowest cost}
  \end{subfigure}
  \caption{Predictions for WS-2}
\end{figure}

\begin{figure}[h!]
\centering
  \begin{subfigure}[b]{0.49\textwidth}
    \includegraphics[trim=0cm 0cm 0cm 2cm,width=1\linewidth]{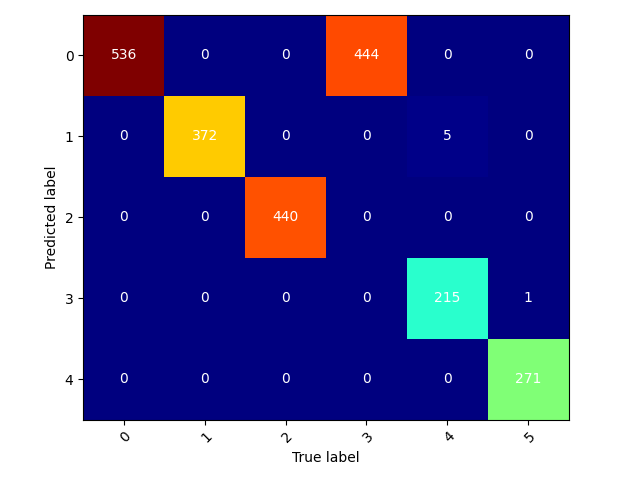}
    \caption{Most common}
  \end{subfigure}
  \begin{subfigure}[b]{0.49\textwidth}
    \includegraphics[trim=0cm 0cm 0cm 2cm,width=1\linewidth]{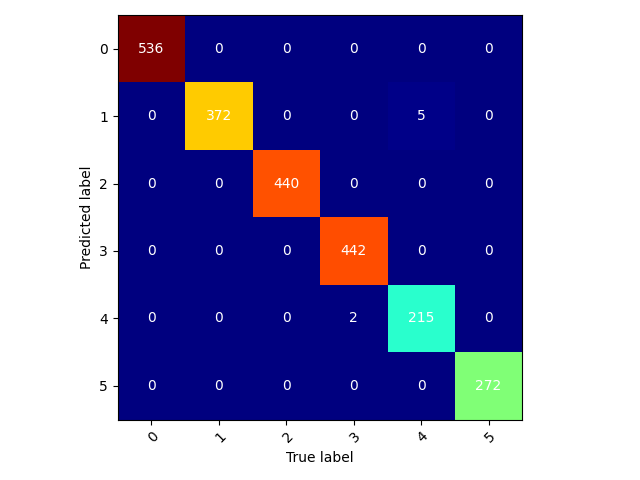}
    \caption{Lowest cost}
  \end{subfigure}
  \caption{Confusion matrices for WS-2}
\end{figure}

\begin{figure}
\begin{floatrow}
\ffigbox{%
      \includegraphics[trim=2cm 1cm 4cm 2cm,clip,scale=0.30]{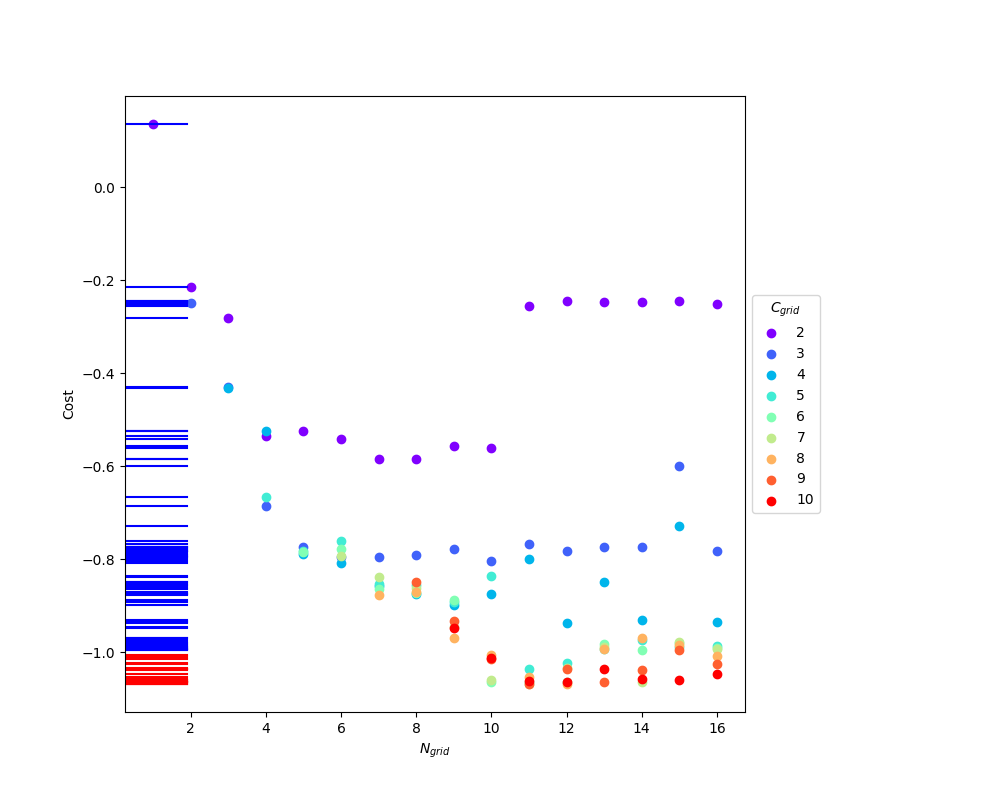}
}{%
           \caption{Grid-search criterion for WS-2-smooth\\ Most common: $N_{grid}=13$, $C_{grid}=6$}
}
\capbtabbox{%

 \begin{tabular}{||c c c c c||} 
 \hline
label &  m & n & $\rho_{w}$ & CEC \\ [0.5ex] 
 \hline\hline
 0 & 1.85 & 1.8 & 0.052 & 0 \\ 
 \hline
 1 & 2.1 & 2.0 & 0.052 & 0 \\
 \hline
 2 & 2.4 & 2.3 & 0.049 & 80 \\
 \hline
 3 & 1.9 & 2.0 & 0.051 & 0 \\
 \hline
 4 & 2.0 & 1.95 & 0.051 & 30 \\
 \hline
 5 & 2.0 & 2.5 & 0.05 & 0 \\ 
 \hline
\end{tabular}

}{%
  \caption{Parameters for WS-2-smooth}%
}
\end{floatrow}
\end{figure}

\begin{figure}[h!]
\centering
  \begin{subfigure}[b]{0.49\textwidth}
    \includegraphics[trim=4cm 2cm 2cm 2cm,clip,width=1\linewidth]{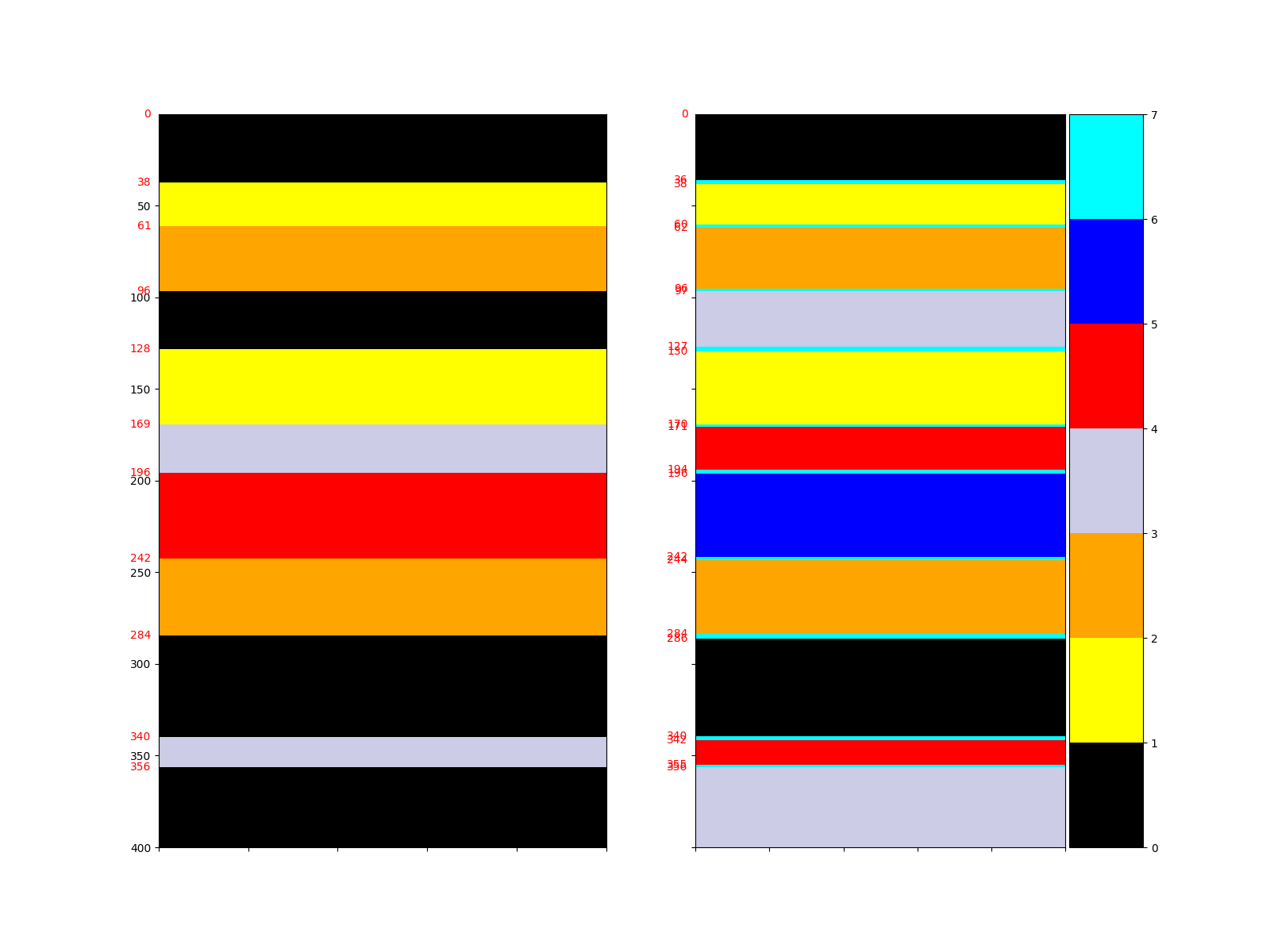}
    \caption{Most common}
  \end{subfigure}
  \begin{subfigure}[b]{0.49\textwidth}
    \includegraphics[trim=4cm 2cm 2cm 2cm,clip,width=1\linewidth]{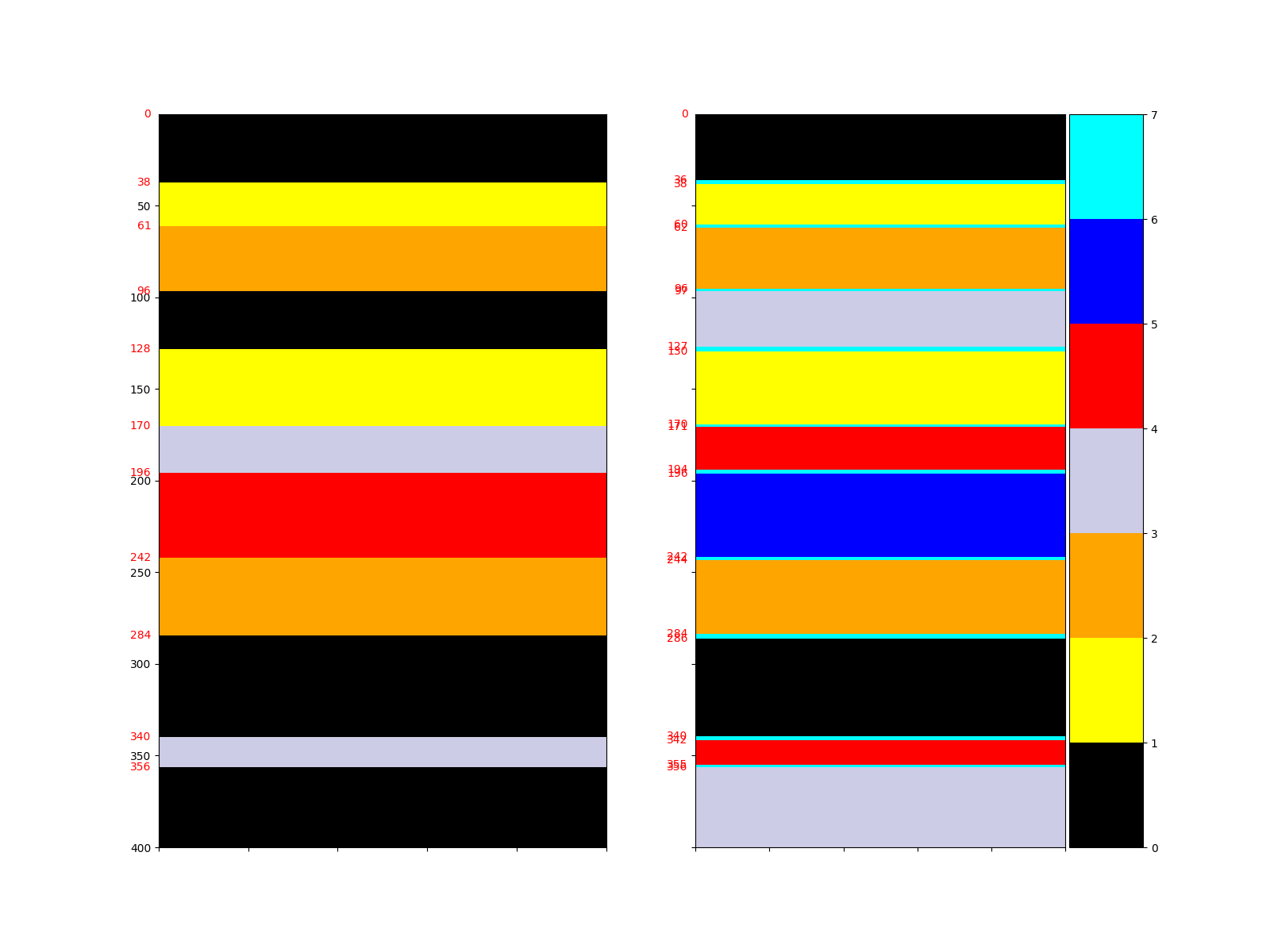}
    \caption{Lowest cost}
  \end{subfigure}
  \caption{Predictions for WS-2-smooth}
\end{figure}

\begin{figure}[h!]
\centering
  \begin{subfigure}[b]{0.49\textwidth}
    \includegraphics[trim=0cm 0cm 0cm 2cm,width=1\linewidth]{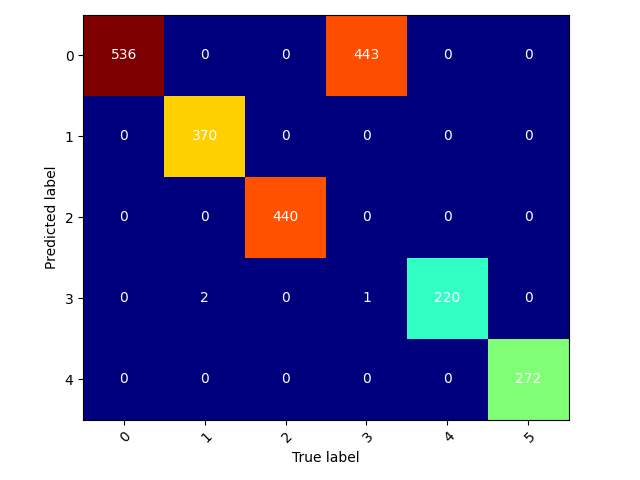}
    \caption{Most common}
  \end{subfigure}
  \begin{subfigure}[b]{0.49\textwidth}
    \includegraphics[trim=0cm 0cm 0cm 2cm,width=1\linewidth]{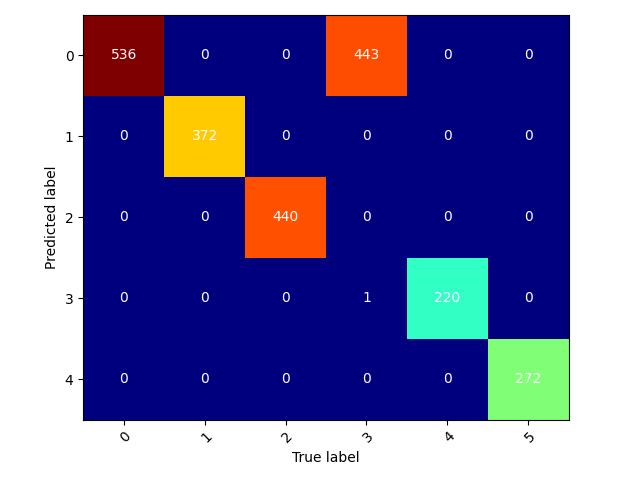}
    \caption{Lowest cost}
  \end{subfigure}
  \caption{Confusion matrices for WS-2-smooth}
\end{figure}

\begin{figure}
	\begin{floatrow}
		\ffigbox{%
			\includegraphics[trim=2cm 1cm 4cm 2cm,clip,scale=0.3]{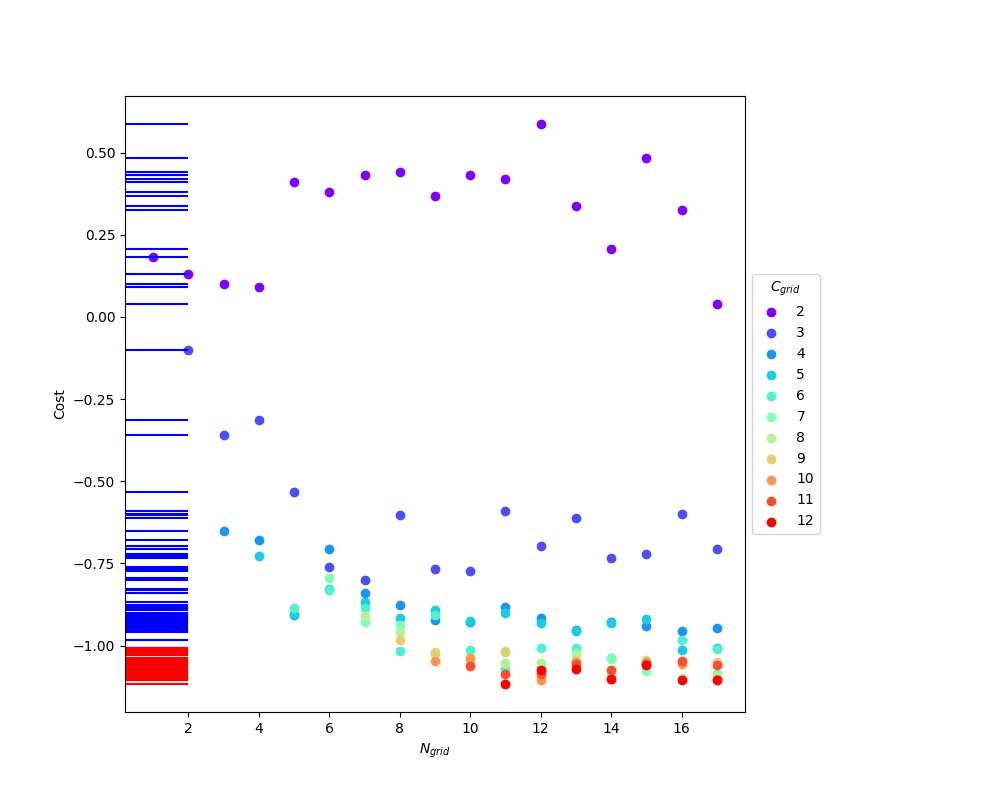}
		}{%
			\caption{Grid-search criterion \\ Most common: $N_{grid}=16$, $C_{grid}=7$}
			\label{fig:grid-WS-3-smooth}%
		}
		\capbtabbox{%
			
			\begin{tabular}{||c c c c c||} 
				\hline
				label & m & n & $\rho_{w}$ & CEC \\ [0.5ex] 
				\hline\hline
				0 & 1.85 & 1.7 & 0.03 & 0 \\ 
				\hline
				1 & 2.0 & 2.0 & 0.03 & 0 \\
				\hline
				2 & 2.05 & 2.0 & 0.029 & 30 \\
				\hline
				3 & 2.3 & 2.1 & 0.031 & 0 \\
				\hline
				4 & 2.5 & 2.2 & 0.049 & 80 \\
				\hline
				5 & 2.0 & 2.5 & 0.05 & 0 \\
				\hline
				6 & 2.0 & 1.9 & 0.05 & 0 \\
				\hline
				7 & 2.1 & 2.1 & 0.051 & 45 \\ 
				\hline
			\end{tabular}
			
		}{%
			\caption{Parameters for WS-3}
			\label{table:WS-3-smooth}%
		}
	\end{floatrow}
\end{figure}

\begin{figure}[h!]
	\centering
	\begin{subfigure}[b]{0.49\textwidth}
		\includegraphics[trim=4cm 2cm 2cm 2cm,clip,width=1\linewidth]{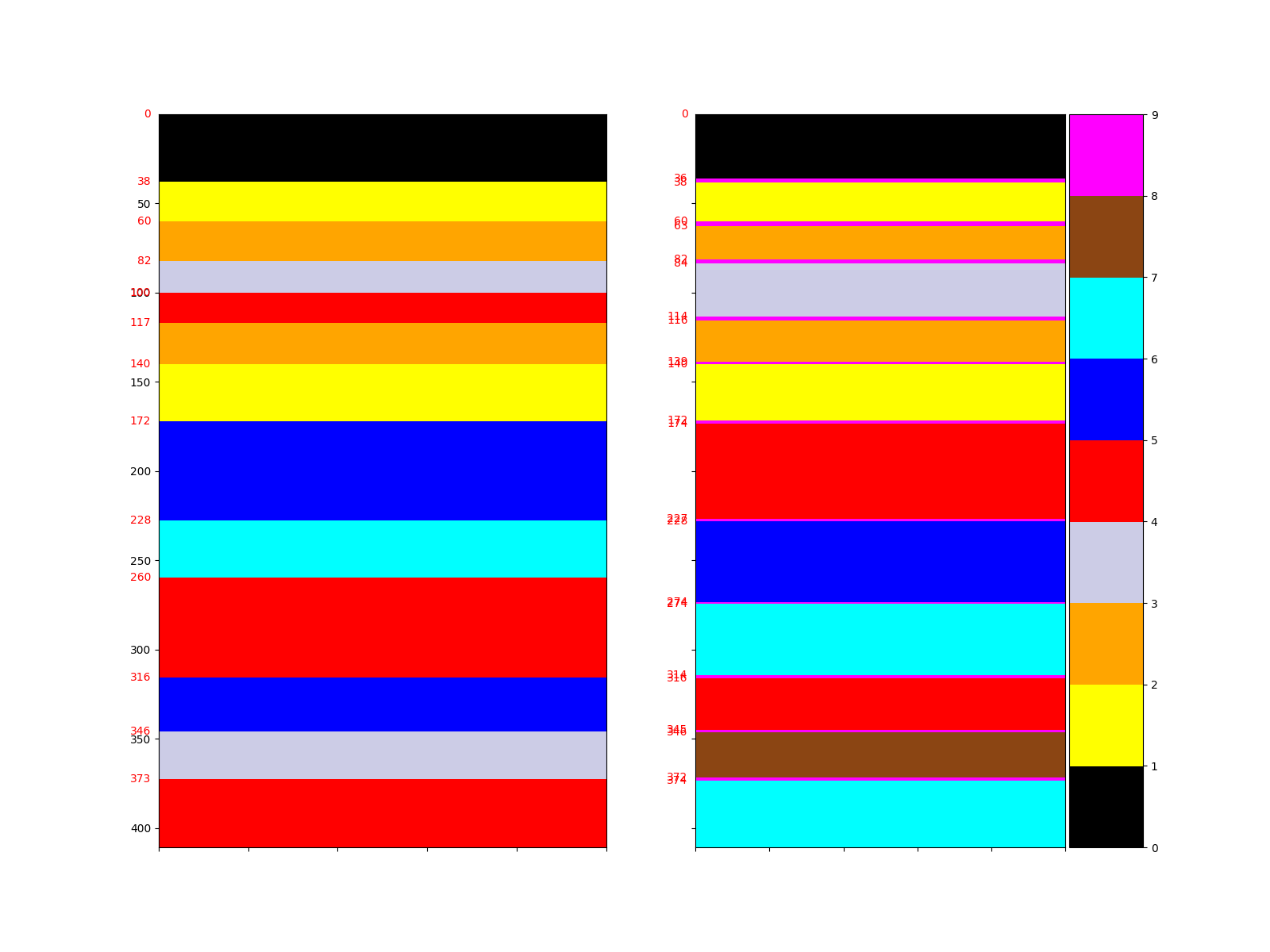}
		\caption{Most common}
	\end{subfigure}
	\begin{subfigure}[b]{0.49\textwidth}
		\includegraphics[trim=4cm 2cm 2cm 2cm,clip,width=1\linewidth]{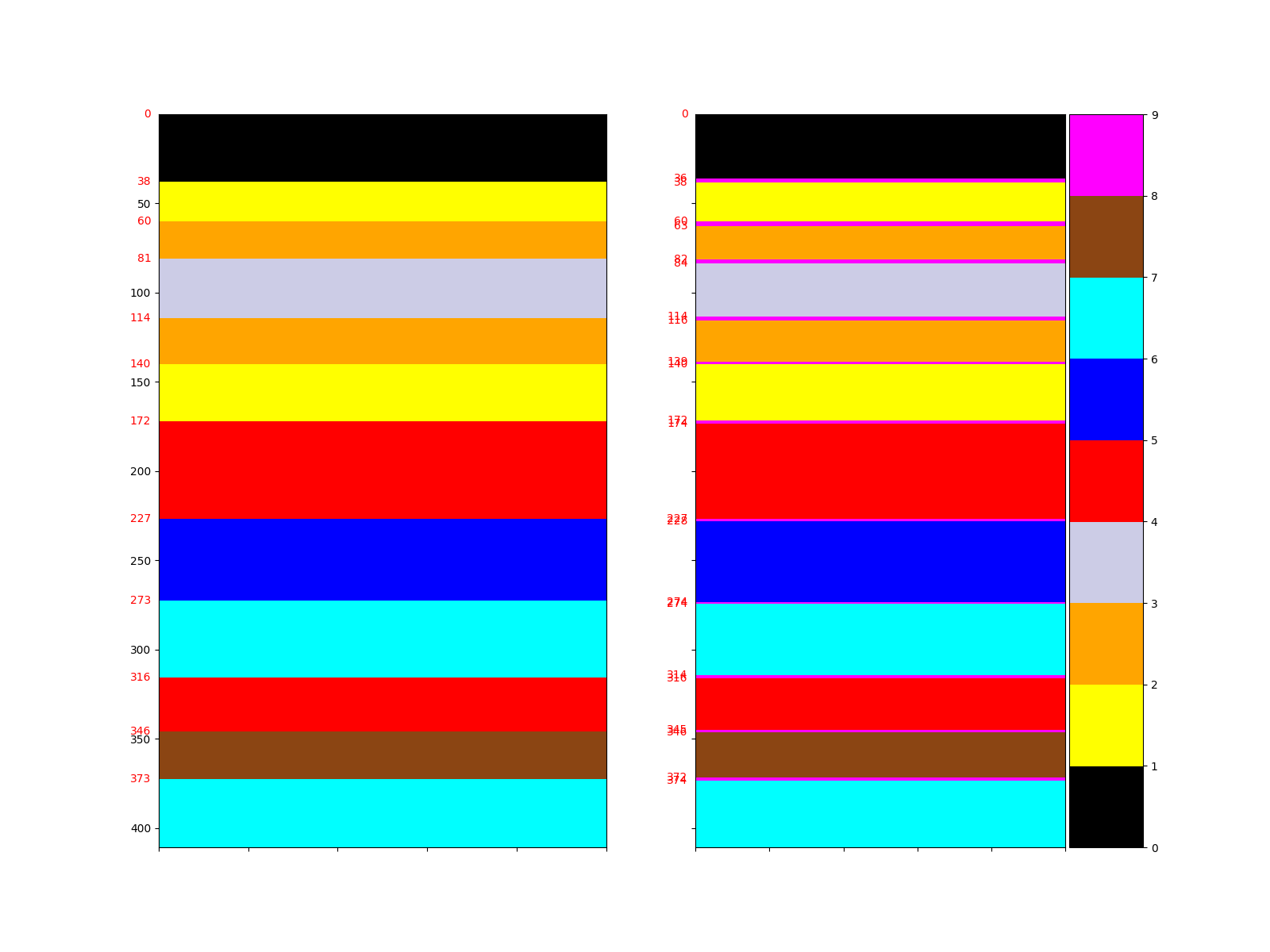}
		\caption{Lowest cost}
	\end{subfigure}
	\caption{Predictions for WS-3-smooth}
\end{figure}

\begin{figure}[h!]
	\centering
	\begin{subfigure}[b]{0.49\textwidth}
		\includegraphics[trim=0cm 0cm 0cm 2cm,width=1\linewidth]{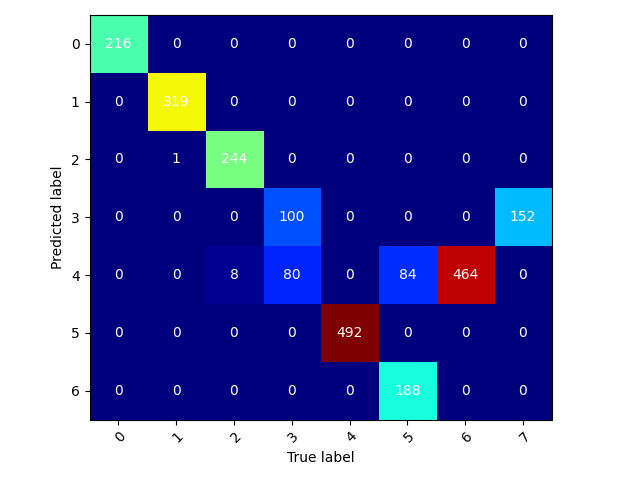}
		\caption{Most common}
	\end{subfigure}
	\begin{subfigure}[b]{0.49\textwidth}
		\includegraphics[trim=0cm 0cm 0cm 2cm,width=1\linewidth]{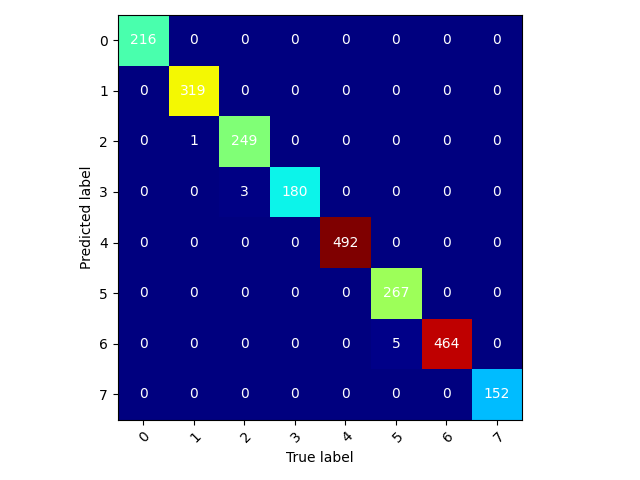}
		\caption{Lowest cost}
	\end{subfigure}
	\caption{Confusion matrices for WS-3-smooth}
\end{figure}

\begin{figure}
\begin{floatrow}
\ffigbox{%
        \includegraphics[trim=2cm 1cm 4cm 2cm,clip,scale=0.30]{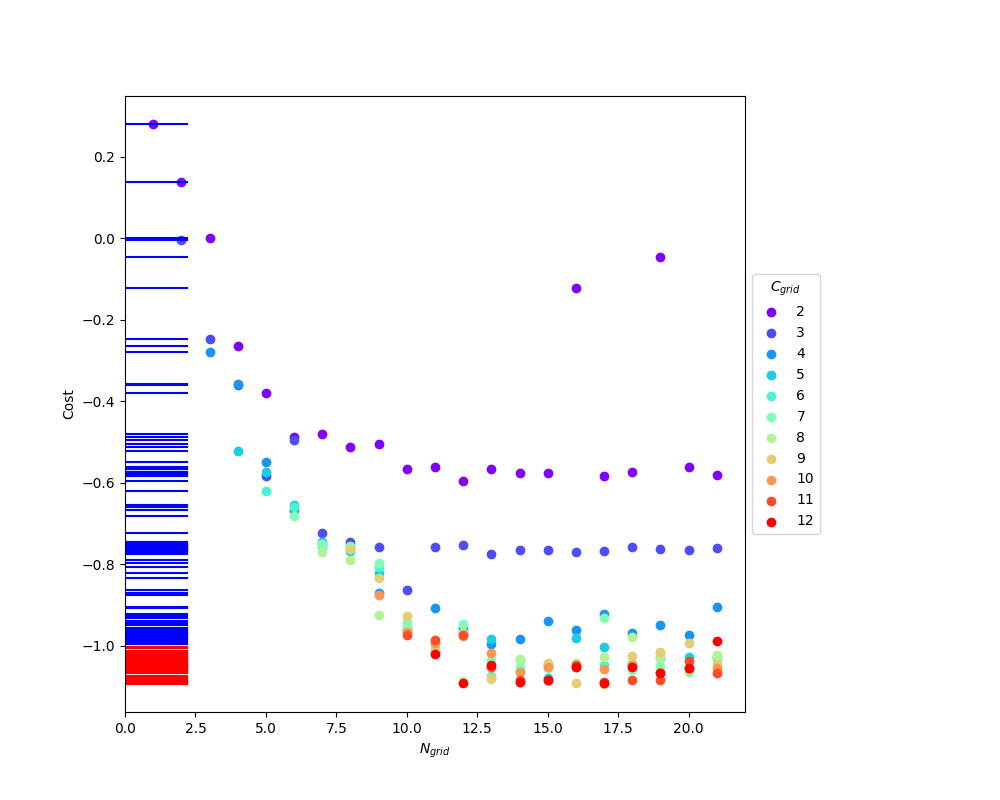}
}{%
   \caption{Grid-search criterion for SGS-1 \\ Most common: $N_{grid}=13$, $C_{grid}=9$}%
}
\capbtabbox{%

 \begin{tabular}{||c c c c c c||} 
 \hline
 label & m & n & $\rho_{w}$ & CEC & Equation\\ [0.5ex] 
 \hline\hline
 0 & 2.0 & 1.95 & 0.051 & 30 & SGS\\ 
 \hline
 1 & 1.85 & 1.8 & 0.052 & 0 & WS \\
 \hline
 2 & 2.1 & 2.0 & 0.052 & 0 & WS \\
 \hline
 3 & 2.4 & 2.3 & 0.049 & 80 & SGS \\
 \hline
 4 & 2.4 & 2.3 & 0.049 & 80 & WS \\
 \hline
 5 & 1.9 & 2.0 & 0.051 & 0 & WS \\
 \hline
 6 & 2.0 & 1.95 & 0.051 & 30 & WS\\
 \hline
 7 & 2.0 & 2.5 & 0.05 & 0 & WS \\ 
 \hline
\end{tabular}

}{%
  \caption{Parameters for SGS-1}%
}
\end{floatrow}
\end{figure}

\begin{figure}[h!]
\centering
  \begin{subfigure}[b]{0.49\textwidth}
    \includegraphics[trim=4cm 2cm 2cm 2cm,clip,width=1\linewidth]{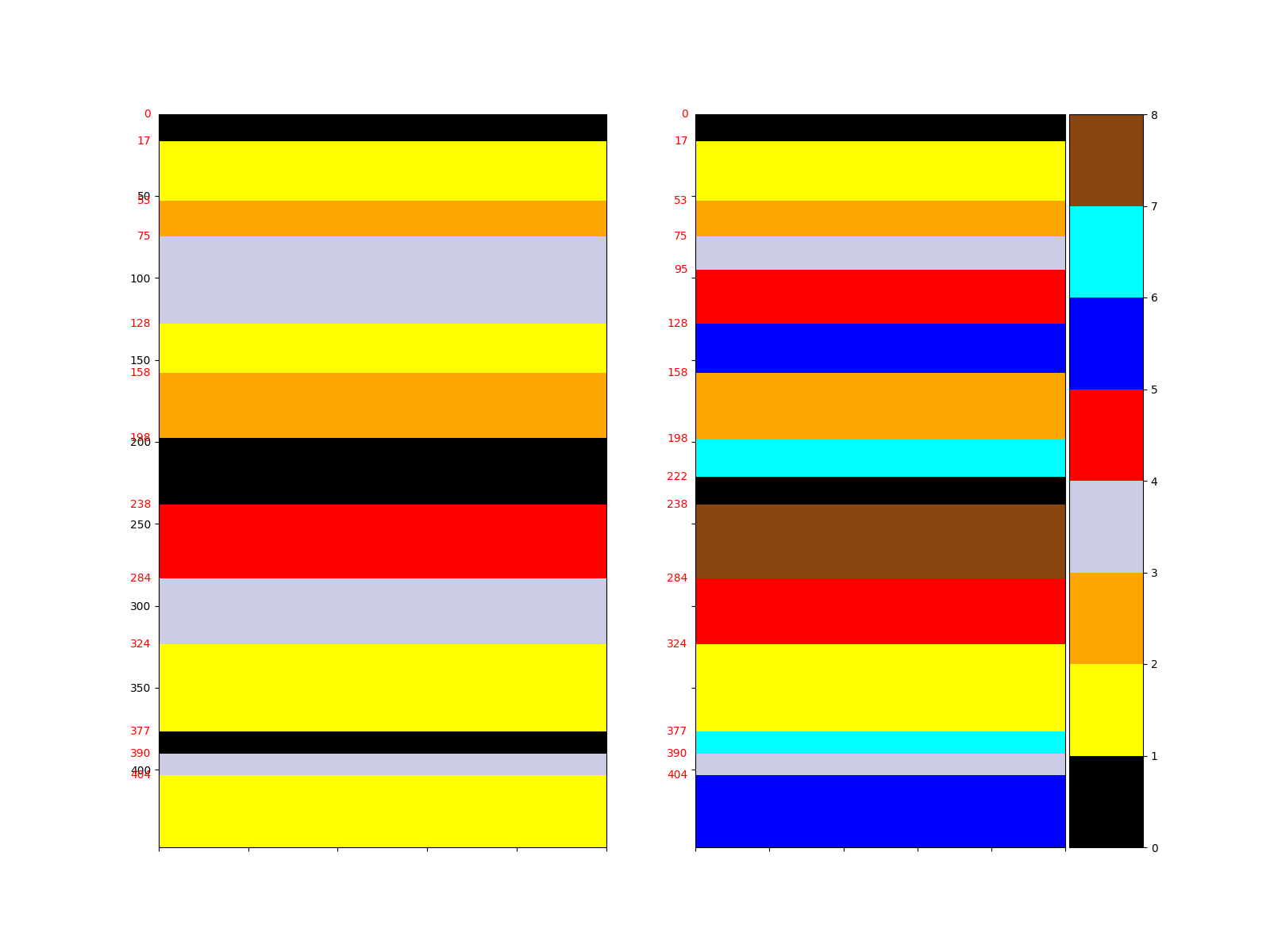}
    \caption{Most common}
  \end{subfigure}
  \begin{subfigure}[b]{0.49\textwidth}
    \includegraphics[trim=4cm 2cm 2cm 2cm,clip,width=1\linewidth]{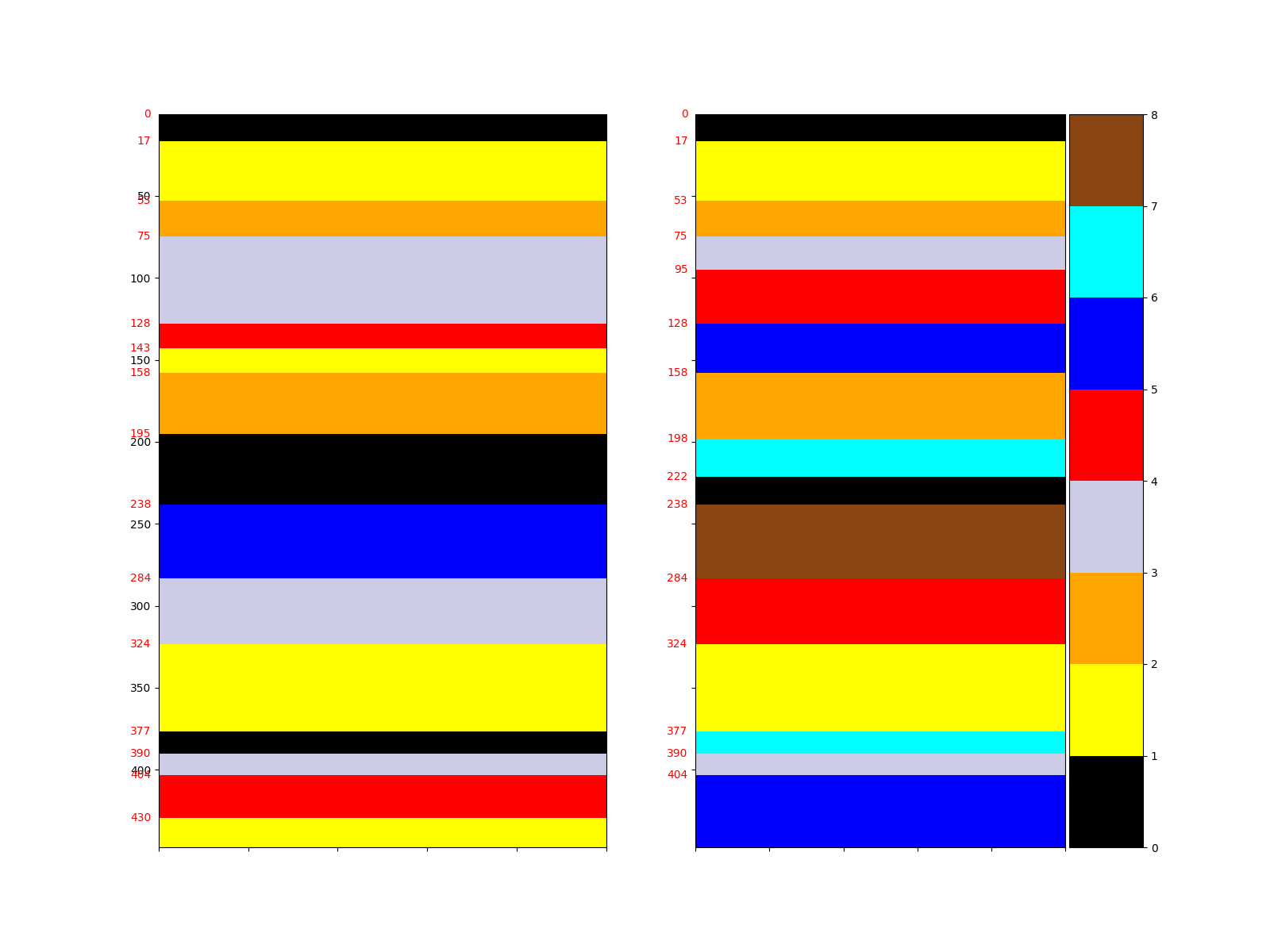}
    \caption{Lowest cost}
  \end{subfigure}
  \caption{Predictions for SGS-1}
\end{figure}

\begin{figure}[h!]
\centering
  \begin{subfigure}[b]{0.49\textwidth}
    \includegraphics[trim=0cm 0cm 0cm 2cm,width=1\linewidth]{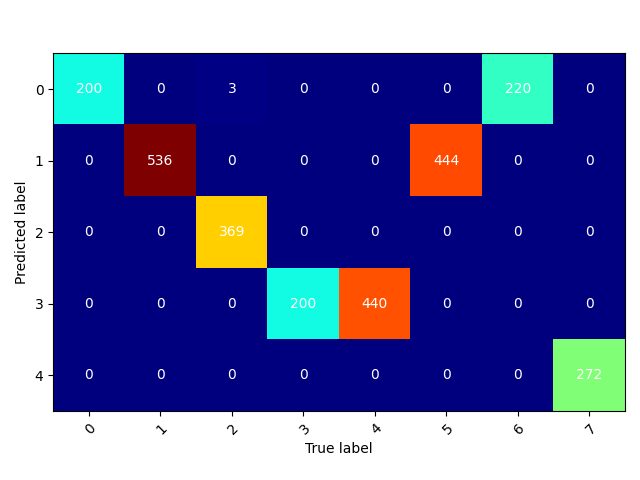}
    \caption{Most common}
  \end{subfigure}
  \begin{subfigure}[b]{0.49\textwidth}
    \includegraphics[trim=0cm 0cm 0cm 2cm,width=1\linewidth]{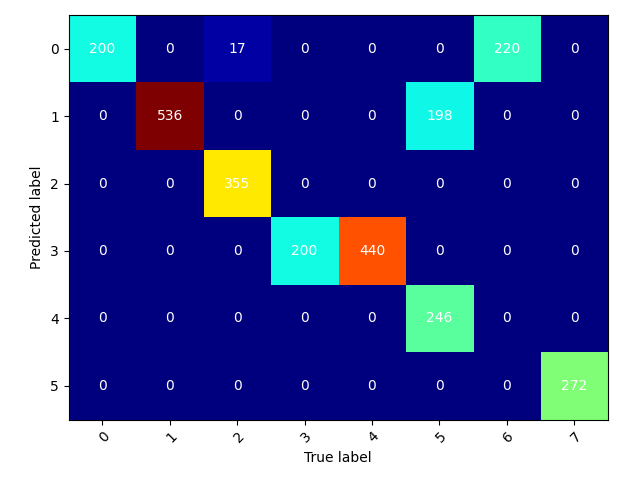}
    \caption{Lowest cost}
  \end{subfigure}
  \caption{Confusion matrices for SGS-1}
\end{figure}

\begin{figure}
\begin{floatrow}
\ffigbox{%
      \includegraphics[trim=2cm 1cm 4cm 2cm,clip,scale=0.30]{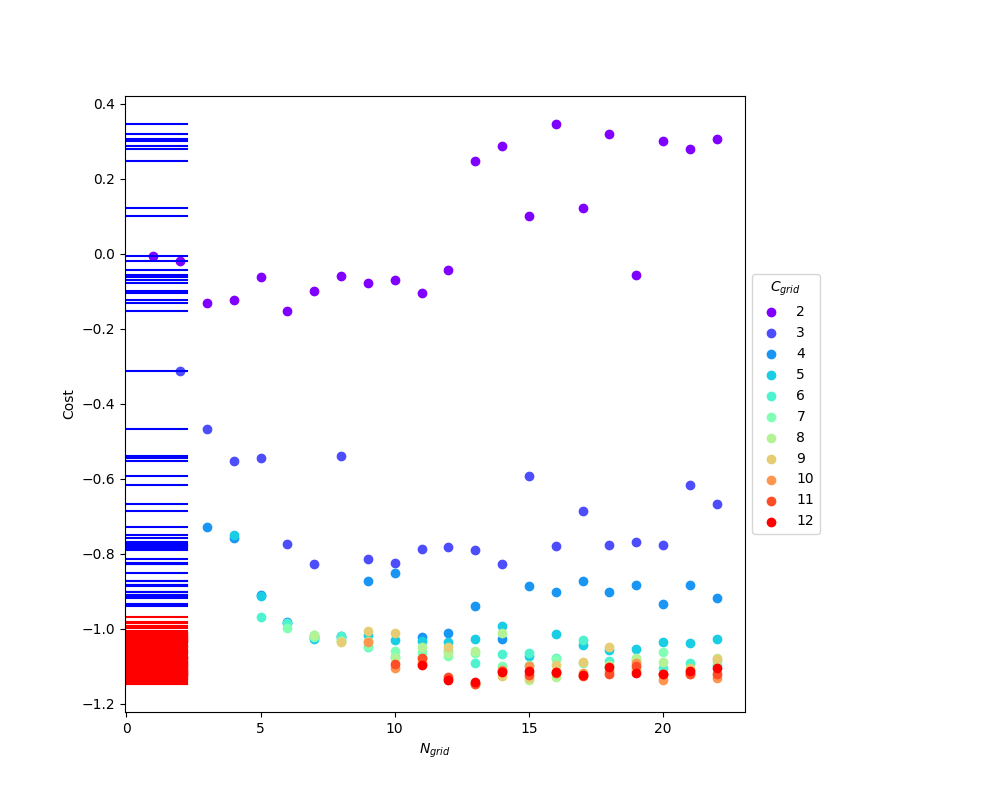}
}{%
           \caption{Grid-search criterion for SGS-2\\ Most common: $N_{grid}=15$, $C_{grid}=10$}
}
\capbtabbox{%

 \begin{tabular}{||c c c c c||} 
 \hline
 label & m & n & $\rho_{w}$ & CEC \\ [0.5ex] 
 \hline\hline
 0 & 1.85 & 1.7 & 0.03 & 0 \\ 
 \hline
 1 & 2.0 & 2.0 & 0.03 & 0 \\
 \hline
 2 & 2.05 & 2.0 & 0.029 & 30 \\
 \hline
 3 & 2.3 & 2.1 & 0.031 & 0 \\
 \hline
 4 & 2.05 & 2.0 & 0.029 & 30 \\
 \hline
 5 & 2.5 & 2.2 & 0.049 & 80 \\
 \hline
 6 & 2.0 & 2.5 & 0.05 & 0 \\
 \hline
 7 & 2.0 & 1.9 & 0.05 & 0 \\ 
 \hline
 8 & 2.5 & 2.2 & 0.049 & 80 \\
 \hline
 9 & 2.1 & 2.1 & 0.051 & 45 \\
 \hline
 10 & 2.1 & 2.1 & 0.051 & 45 \\ 
 \hline
\end{tabular}

}{%
  \caption{Parameters for SGS-2}%
}
\end{floatrow}
\end{figure}

\begin{figure}[h!]
\centering
  \begin{subfigure}[b]{0.49\textwidth}
    \includegraphics[trim=4cm 2cm 2cm 2cm,clip,width=1\linewidth]{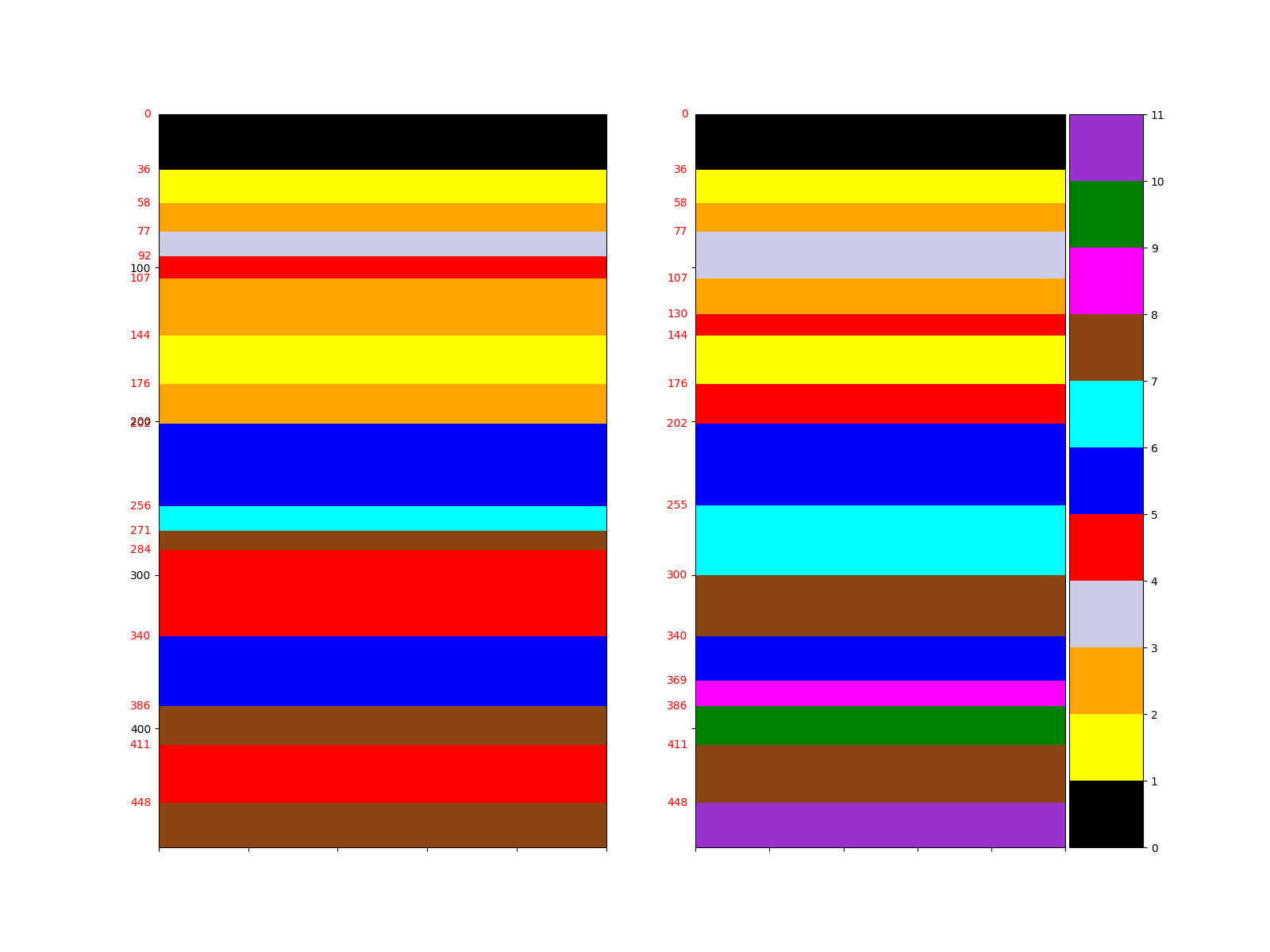}
    \caption{Most common}
  \end{subfigure}
  \begin{subfigure}[b]{0.49\textwidth}
    \includegraphics[trim=4cm 2cm 2cm 2cm,clip,width=1\linewidth]{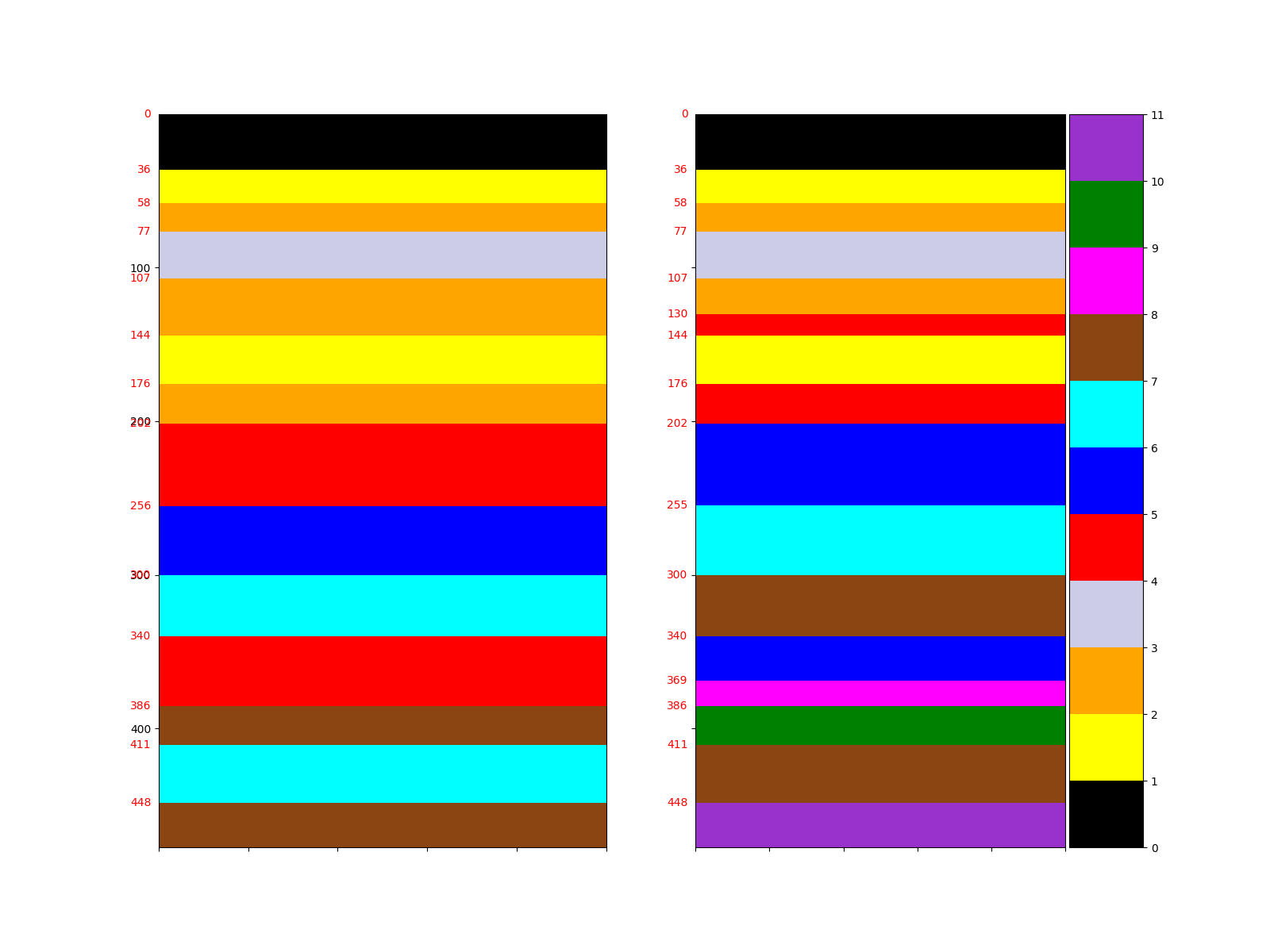}
    \caption{Lowest cost}
  \end{subfigure}
  \caption{Predictions for SGS-2}
\end{figure}

\begin{figure}[h!]
\centering
  \begin{subfigure}[b]{0.49\textwidth}
    \includegraphics[trim=0cm 0cm 0cm 2cm,width=1\linewidth]{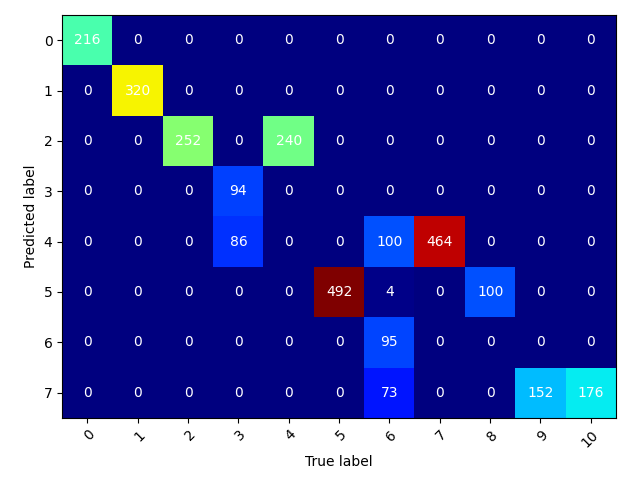}
    \caption{Most common}
  \end{subfigure}
  \begin{subfigure}[b]{0.49\textwidth}
    \includegraphics[trim=0cm 0cm 0cm 2cm,width=1\linewidth]{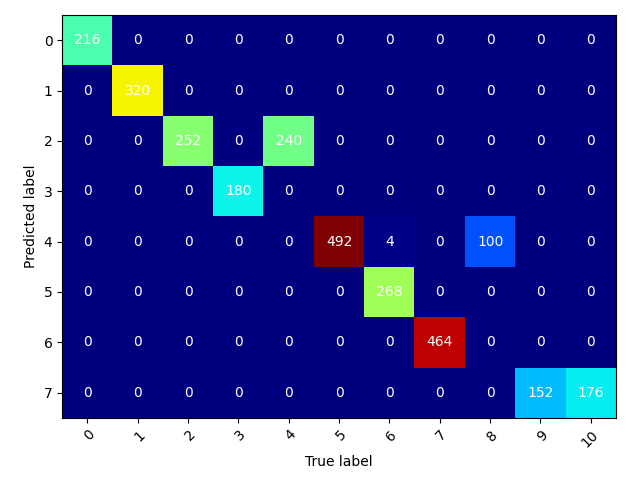}
    \caption{Lowest cost}
  \end{subfigure}
  \caption{Confusion matrices for SGS-2}
\end{figure}
\end{appendices}
\end{document}